\pdfoutput=1

\documentclass[11pt]{article}

\usepackage[]{EMNLP2022}

\usepackage{times}
\usepackage{latexsym}

\usepackage[T1]{fontenc}

\usepackage[utf8]{inputenc}

\usepackage{microtype}

\usepackage{inconsolata}

\usepackage{microtype}
\usepackage{microtype}
\usepackage{amsmath,amssymb}
\usepackage{epsfig,graphicx,subfigure,caption}
\usepackage{algpseudocode}
\usepackage[normalem]{ulem}
\usepackage{amsmath}
\usepackage[linesnumbered,algoruled,boxed,noend]{algorithm2e}
\usepackage{xcolor}
\usepackage{color, colortbl}
\usepackage{multirow,booktabs, hhline}
\usepackage{amsmath, bm}
\usepackage[linesnumbered,algoruled,boxed,noend]{algorithm2e}
\usepackage{wrapfig}

\useunder{\uline}{\ul}{}
\definecolor{Gray}{gray}{0.9}
\definecolor{Blue}{rgb}{0.8,0.85,1}
\definecolor{Red}{rgb}{1,0.85,0.8}
\newcommand{\MyColorBox}[2][red]%
{%
    \settowidth{\Width}{#2}%
    \colorbox{#1}%
    {%
        \raisebox{-\DepthReference}%
        {%
                \parbox[b][\HeightReference+\DepthReference][c]{\Width}{\centering#2}%
        }%
    }%
}

\def\autotrigger{\textsc{AutoTriggER}}

\title{AutoTriggER: Label-Efficient and Robust Named Entity Recognition \\ with Auxiliary Trigger Extraction}

\author{
Dong-Ho Lee\textsuperscript{1,4}\thanks{~~{The first two authors contributed equally.}},~
Ravi Kiran Selvam\textsuperscript{1}$^*$,~
Sheikh Muhammad Sarwar\textsuperscript{3}\thanks{work done before joining Amazon.com.},~
Bill Yuchen Lin\textsuperscript{1},\\ 
\textbf{
{Fred Morstatter}\textsuperscript{4},~
Jay Pujara\textsuperscript{4},~
{Elizabeth Boschee}\textsuperscript{4},~
{James Allan}\textsuperscript{2},~
{Xiang Ren}\textsuperscript{1,4}
}
\\
\textsuperscript{1}Department of Computer Science, University of Southern California\\
\textsuperscript{2}CIIR, Manning CICS, University of Massachusetts Amherst, \textsuperscript{3}Amazon.com\\
\textsuperscript{4}Information Science Institute, University of Southern California\\

{\texttt{\{dongho.lee,rselvam,yuchen.lin,xiangren\}@usc.edu},} \\
\texttt{allan@cs.umass.edu}, \texttt{smsarwar@amazon.com}, \texttt{\{fredmors,jpujara,boschee\}@isi.edu}
}

\begin{document}
\maketitle
\begin{abstract}
Deep neural models for named entity recognition (NER) have shown impressive results in overcoming label scarcity and generalizing to unseen entities by leveraging distant supervision and auxiliary information such as explanations.
However, the costs of acquiring such additional information are generally prohibitive.
In this paper, we present a novel two-stage framework (\textsc{AutoTriggER}) to improve NER performance by automatically generating and leveraging ``\textit{entity triggers}'' which are human-readable cues in the text that help guide the model to make better decisions.
Our framework leverages post-hoc explanation to generate rationales and strengthens a model's prior knowledge using an embedding interpolation technique. This approach allows models to exploit triggers to infer entity boundaries and types instead of solely memorizing the entity words themselves.
Through experiments on three well-studied NER datasets, \textsc{AutoTriggER} shows strong label-efficiency, is capable of generalizing to unseen entities, and outperforms the RoBERTa-CRF baseline by nearly 0.5 F1 points on average.
\end{abstract}
\section{Introduction} 



\begin{figure}[t!]
    \centering 
    \includegraphics[scale=0.6]{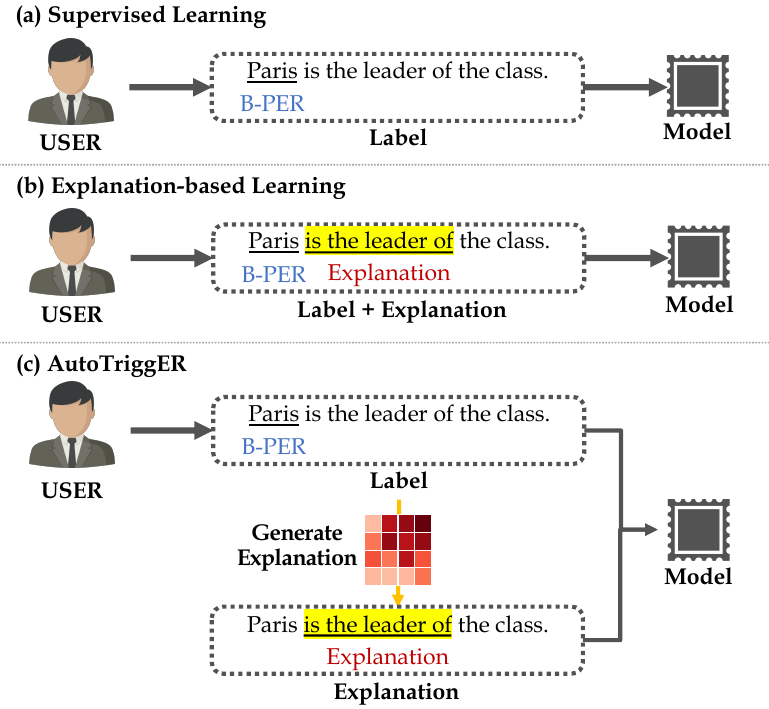}
    \caption{Existing explanation-based learning frameworks mostly rely on humans provided labeling explanations while our framework automatically generates and leverages explanations to NER.}
    \label{fig:overview}
\end{figure}




Named Entity Recognition (NER) serves as a key building block in information extraction systems.
Recent advances in deep neural models for NER have yielded state-of-the-art performance when sufficient human annotations are available~\citep{Lample2016NeuralAF, liu2017empower, Peters2017SemisupervisedST, Ma2016EndtoendSL}.
However, such success cannot easily transfer to practitioners developing NER systems in specific domains (\textit{e.g.}, biomedical papers, financial reports, legal documents),
where domain-expert annotations are expensive and slow to obtain.
Moreover, NER systems face challenges in real-world applications where novel entities, unseen in training data, are often encountered~\cite{lin-etal-2021-rockner}.
Recent attempts addressing label scarcity and improving generalization to unseen entities have explored various types of human-curated resources as auxiliary supervision, such as entity dictionaries~\citep{Peng2019DistantlySN, autoner, yangner, Liu2019TowardsIN}, labeling rules~\citep{safranchik:aaai20, jiang2020cold}, prompts~\cite{ding2021prompt,cui2021template}, demonstrations~\cite{lee-etal-2022-good}, retrieved context~\cite{wang-etal-2021-improving}, and labeling explanations~\citep{Hancock2018TrainingCW, Wang2020Learning,Ye2020TeachingMC, lin-etal-2020-triggerner, lee-etal-2020-lean}.
In particular, human-provided labeling explanations as auxiliary supervision signals are cost-effective compared to collecting label-only annotations for larger number of instances.
For the NER task, ~\citet{lin-etal-2020-triggerner} introduced the concept of an \textit{entity trigger}, an effective way to represent explanations for the labeling decisions (See Figure~\ref{fig:overview} (a) vs. (b)).



However, prior works have the following limitations:
(1) \textit{Expense of collecting rationales}: 
Prior works primarily use a limited number of \textit{crowd-soured} explanations (e.g., \textit{entity trigger}) for improving data (label) efficiency of model training.
While such human-curated auxiliary supervision is of high quality, the crowd-sourcing procedure can be very expensive and time-consuming.
This largely limits the scale and domains of the collected rationales;
(2) \textit{Unseen entity generalization}:
The biggest advantage of leveraging such explanations for NER is a generalization ability towards unseen entities.
Reinforcing this prior knowledge in the model rather than memorizing the entity words themselves can make the model robust against unseen entities during training.
However, prior works do not evaluate such generalization ability.

To address these limitations, we propose a two-stage NER framework, named \autotrigger~, which improves label efficiency and model robustness without human efforts (see Figure~\ref{fig:overview} (c)).
First, it \textit{automatically} generates entity triggers as explainable features to reduce human effort on collecting rationales.
It then strengthens the prior knowledge in the model by interpolating model embeddings of entity-masked sentence which force the model to rely more on triggers and trigger-masked sentence which force the model to rely more on the entity.

The \textit{first} stage of our framework (Sec.~\ref{ssec:soc}) aims to \textit{automatically extract entity triggers} using a post-hoc explanation.
We propose to exploit the syntactic features of a sentence for assigning importance scores to a group of input tokens such that we can extract entity triggers as auxiliary supervision.
Specifically, we form a collection of trigger candidates for an entity in a sentence using the phrases extracted from its \textit{constituency parsing tree}~\citep{joshi-etal-2018-extending}.
Then we score each trigger candidate based on its ability to predict the target entity, while varying the surrounding context to ensure trigger robustness.



The \textit{second} stage (Sec.~\ref{ssec:tin}) focuses on how to use triggers, which are useful contextual clues, in making the prediction.
We propose \textit{Trigger Interpolation Network} (TIN), a novel architecture that effectively uses trigger-labeled NER data to train a model.
Here, we employ two separate masking passes when learning our model's embeddings: one masking the entity words (forcing the model to rely more on the triggers) and one masking the triggers (forcing the model to rely more on the entity words).
We then interpolate the embeddings of both entity-masked and trigger-masked sentences to learn a mixed sentence representation, which is provided as input to a standard sequence labeling task.
In this manner, the TIN can effectively learn to focus on useful contextual clues to infer entity boundaries and types with contextualized embeddings from pre-trained language models such as BERT~\citep{devlin-etal-2019-bert} while it may not miss the entities whose type can be determined by itself. (e.g., inputs that contain only the entities).

Extensive experimental results on several domains show that \autotrigger~ framework consistently outperforms baseline methods by 0.5 F1 points on average in fully supervised setting.
Our work shows strong performance especially in label scarcity setting and unseen entity generalization.
In the label scarcity setting ranging from 50 to 200 number of train instances, assuming a task that needs annotation from scratch, our model gains more than 3-4 F1 score on average.
Also, for the filtered test set where the entities did not appear in the train and development sets, assuming a real-world applications in which entities can be out of the distribution of the train data, our model gains more than 2-3 F1 score on average.

\section{Problem Formulation}
\label{sec:problem}


\smallskip
\noindent
\textbf{Named Entity Recognition.}\quad 
We let $\mathbf{x} = [x^{(1)}, x^{(2)}, \dots x^{(n)}]$ denote the sentence consisting of a sequence of $n$ words and $\mathbf{y} = [y^{(1)}, y^{(2)}, \dots y^{(n)}]$ denote the NER-tag sequence.
The task is to predict the entity tag $y^{(i)}\in\mathcal{Y}$ for each word $x^{(i)}$, where $\mathcal{Y}$ is a pre-defined set of tags such as \{\textsc{B-PER}, \textsc{I-PER}, \dots, \textsc{O}\}.
We let $\mathcal{D}_{L}$ denote the labeled dataset consisting of the set of instances $\{(\mathbf{x_i}, \mathbf{y_i})\}$, where $\mathbf{x_i}$ is the $i$-th input sentence and $\mathbf{y_i}$ is its output tag sequence.
Here, we use BIOES scheme~\cite{chiu-nichols-2016-named}.

\begin{figure}[t]
\vspace{-0.3cm}
    \centering 
    \includegraphics[scale=0.55]{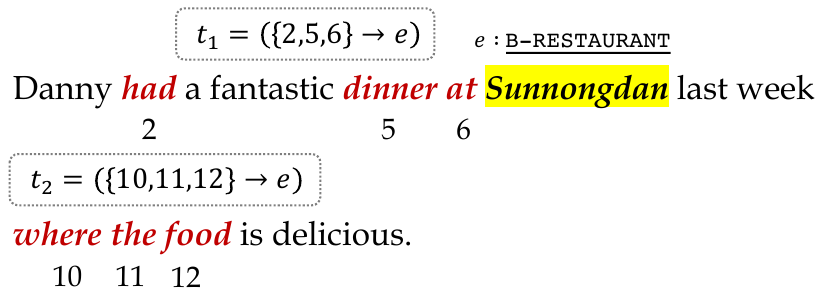}
    \caption{\textbf{Example of entity trigger.} Entity trigger $t_i$ is a cue phrase toward the entity $e$ in the sentence, which is represented by a set of corresponding word indices. Both entity triggers ($t_1$,$t_2$) are associated to the same entity $e$ (``\textit{Sunnongdan}'') typed as restaurant.}
        \label{fig:trigexample}
    \vspace{-0.3cm}
\end{figure}

\paragraph{Entity Trigger} 
Extending from extractive rationales for text classification~\cite{deyoung-etal-2020-eraser,jain-etal-2020-learning}, an \textit{entity trigger} is a form of extractive explanatory annotation for NER, defined as \textit{a group of words that help explain the recognition process of an entity mentioned in the sentence}~\cite{lin-etal-2020-triggerner}. For example, in Figure~\ref{fig:trigexample},
``\textit{had ... dinner at}''
and ``\textit{where the
food}'' are two distinct triggers associated with the
\textsc{Restaurant} entity ``\textit{Sunnongdan}".
Formally, given a labeled NER instance $(\mathbf{x},\mathbf{y})$,
we use $T$ to denote the set of entity triggers for this instance. 
Each trigger $t_{i}\in T$ is associated with an entity $e$ and a set of word indices $\{w_{i}\}$.
That is, $t = (\{w_1, w_2, \dots \} \rightarrow e)$ represents an entity trigger. For example, in Figure~\ref{fig:trigexample}, $t_1=\{2,5,6\}$ is an entity trigger identified for entity ``\textit{Sunnongdan}" in the sentence. 
A \textit{trigger-labeled NER dataset}, $\mathcal{D}_{T}= \{(\mathbf{x_i}, \mathbf{y_i}, T(\mathbf{x_i},\mathbf{y_i}))\}$, consists of examples in a labeled NER dataset $\mathcal{D}_L$ with their associated entity triggers.


\paragraph{Problem Definition} 
We focus on the problem of how to \textit{automatically} extract entity triggers to create a trigger-labeled dataset $\mathcal{D}_{T}$ from $\mathcal{D}_{L}$ without manual effort and to cost-effectively train a NER model using $\mathcal{D}_{T}$.
Here, we aim to achieve better performance and robustness over existing NER models using the same amount of training instances.
Moreover, to check the quality of the automatically extracted triggers, we compare with the model using human-labeled triggers.

\section{Approach}
\label{sec:framework}



\textsc{AutoTriggER} is a two-stage architecture that begins with an \textit{automatic trigger extraction} stage followed by a \textit{trigger interpolation network} (\texttt{TIN}). It automatically extracts and scores the importance of entity trigger phrases in the first stage (Sec.~\ref{ssec:soc}) and uses them in the later stage to train the NER model (Sec.~\ref{ssec:tin}).


\begin{figure}[t]
\vspace{-0.6cm}
    \centering 
    \includegraphics[scale=0.54]{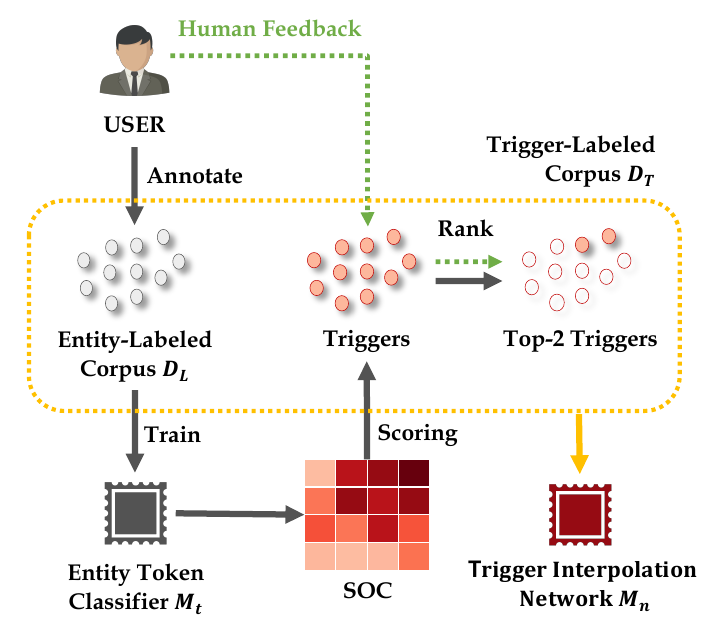}
    \caption{\textbf{Overview of the proposed \autotrigger.} It trains an entity-token classifier $\mathcal{M}_t$ with entity-labeled corpus $\mathcal{D}_L$ and uses the sampling-and-occlusion (SOC) algorithm to extract triggers. There is a provision for leveraging human feedback in the framework for refining automatically generated triggers. Trigger Interpolation Network (TIN) learns the NER model from the trigger-labeled corpus.
    }
    \label{fig:framework}
\end{figure}

\subsection{Automatic Trigger Extraction}
\label{ssec:soc}

\textit{Automatic trigger extraction} is the first stage of our \textsc{AutoTriggER} framework.
Different from prior works on extracting rationales and explanations for sentence classification~\cite{Ribeiro2016WhySI,li2016understanding}, we look to extract triggers for explaining the detection (occurrence) of an entity in the sentence. 
In our study, we extend the \textit{sampling and occlusion} (\texttt{SOC}) algorithm~\citep{jin2020towards}, a feature attribution algorithm for post-hoc model explanation, to extract entity triggers. 
Here, we exploit a token classifier to model the score function of the target entity in the sentence and limit the search space to a set of phrases from the constituent parse tree.
Specifically, given a labeled NER corpus $(\mathbf{x_i},\mathbf{y_i})\in\mathcal{D}_{L}$, we consider four main steps for entity trigger extraction: 1) generating a set of candidate phrases $\mathcal{P}$, 2) constructing entity token classifier $\mathcal{M}_t$, 3) trigger phrase scoring, and 4) trigger phase selection. 

\begin{figure}[t]
    \centering 
    \includegraphics[scale=0.31]{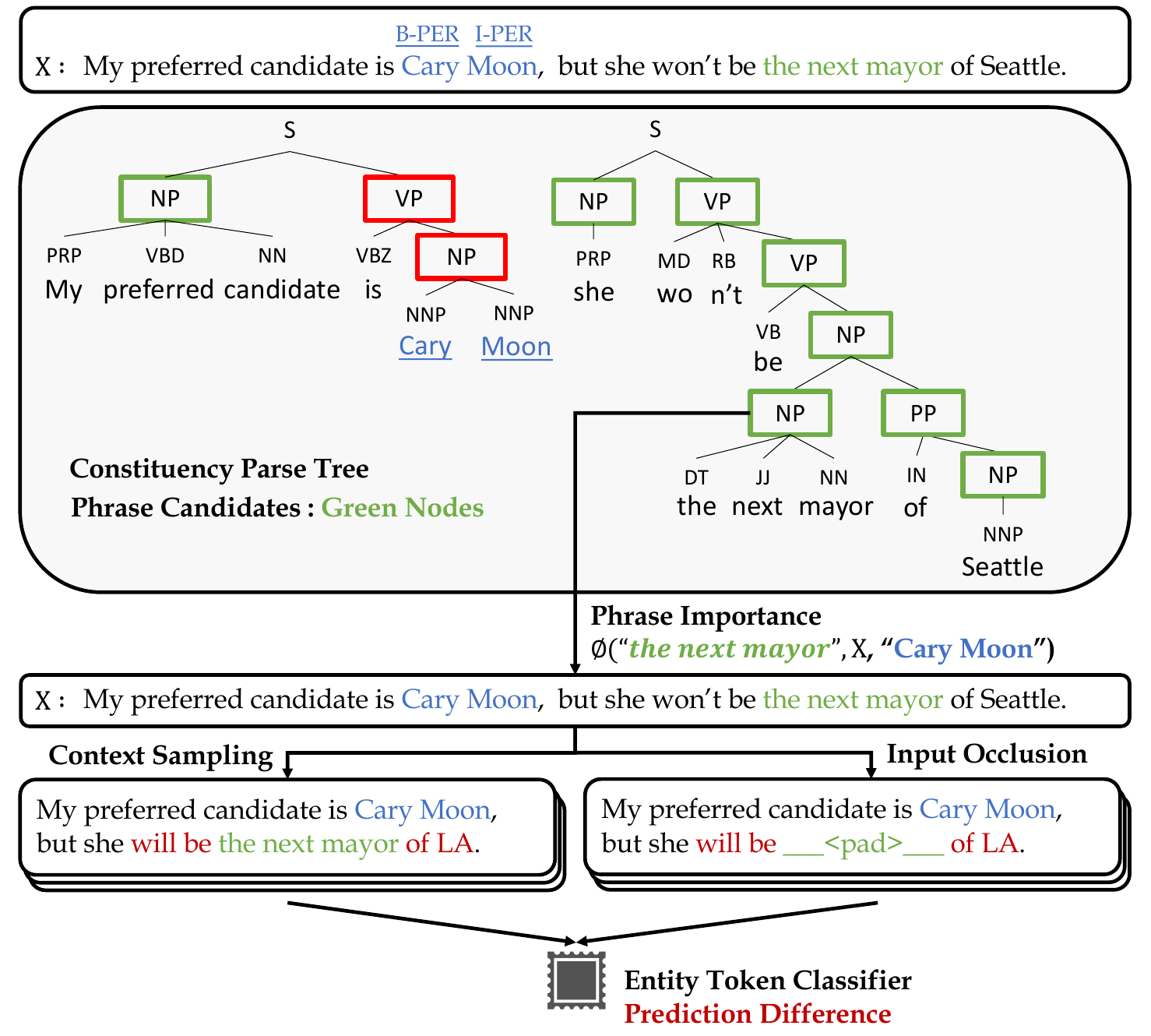}
    \caption{\textbf{Overview of the Sampling and Occlusion (\textsc{SOC})}. It creates a set of phrase candidates with phrase nodes of the constituency parse tree, and then computes the phrase importance by average prediction difference between context sampled sentences and its phrase-masked sentences. Note that the entity mention ``\textit{Cary Moon}'' is not included as candidate. 
    }
    \label{fig:soc}
\end{figure}

\paragraph{Candidate Generation}  
Given a labeled sentence $(\mathbf{x_i},\mathbf{y_i})$, we obtain its constituency parse tree and consider the set of phrase nodes $\mathcal{P}$ from the tree as trigger candidates (see Figure.~\ref{fig:soc} for an illustration with examples). 
\texttt{SOC} aims to compute context-independent phrase-level importance for sequence classification tasks such as sentiment analysis and relation extraction~\cite{jin2020towards}.
It uses agglomerative clustering~\cite{singh2018hierarchical} to effectively compute the phrase-level importance without evaluating all the possible phrases in the sentence.
Here, we provide pre-defined hierarchy for limiting the search space to a set of phrases from the constituent parse tree.
Specifically, given a sentence instance $\mathbf{x_i}\in\mathcal{D}_{L}$ and a target entity mentioned in this sentence $e\in\mathbf{x_i}$, we generate a set of phrase candidate $\mathcal{P}=\{\mathbf{p_i}\}$, where $\mathbf{p_i} = (w_s,w_e)$. Here $(w_s,w_e)$ denote the start and end index of the phrase span $\mathbf{p_i}$.
To generate $\mathcal{P}$, we parse the input sentence $\mathbf{x_i}$ using an off-the-shelf constituency parser~\footnote{https://stanfordnlp.github.io/CoreNLP/parse.html}
and collect phrase nodes in the parse tree as the set of candidate phrases $\mathcal{P}$.
To avoid including target entity as an entity trigger, we discard phrases that contain the target entity, i.e., $\{\mathbf{p_j}|e\in\mathbf{p_j}\}$.

\paragraph{Entity Token Classifier Construction}
In order to apply \texttt{SOC} to the sequence labeling task, we propose a way to derive the importance score of the phrase $\mathbf{p}$ specific to the input sentence $x$ and target entity $e$.
To achieve this, we use a token classification model $\mathcal{M}_t$ to generate prediction scores in the label space (e.g., probability for target class of a token).
Given an input sentence $\mathbf{x_i} = [x^{(1)}_i, x^{(2)}_i, \dots x^{(n)}_i]$, $\mathcal{M}_t$ classifies each token $x^{(j)}_i$ to the named entity tag $y^{(j)}_i \in \mathcal{Y}$ where $\mathcal{Y}$ is a predefined set of named entity tags such as B-PER, I-PER and O.
We learn $\mathcal{M}_t$ with labeled corpus $\mathcal{D}_{L}$ to fit a probability distribution $\mathbb{P}(\mathbf{y}|\mathbf{x})$.
Then, we can derive the prediction score function $s$ towards the target entity $e$ which is computed as the average conditional probability over tokens of the target entity $x^{(j)}\in e$ as follows:
\begin{equation}
    \label{eq:seq_score}
    s(\mathbf{x}, e) = \frac{1}{|e|}\sum_{x^{(j)}\in e} \mathbb{P}(\mathbf{y^{(j)}}|x^{(j)}).
\end{equation}


\noindent
\textbf{Phrase Scoring.}\quad 
Once we have the set of candidate phrases $\mathcal{P}$ for the given input instance $(x_i,y_i)$, we aim to measure the score of each of the candidate phrase $\mathbf{p}$ w.r.t the target entity $e$.
We use the score function $s$ of the $\mathcal{M}_t$ to measure the importance score of each phrase $\mathbf{p}$.
Here, we have two steps: (1) \textit{input occlusion}, (2) \textit{context sampling}.

\textit{Input occlusion}~\citep{li2016understanding} computes the importance of $\mathbf{p}$ specific to the entity $e$ in the input $\mathbf{x}$ by measuring the prediction difference caused by replacing the phrase $\mathbf{p}$ with padding tokens $0_\mathbf{p}$.
We use $s\left(\mathbf{x}_{-\mathbf{p}}, e ; \mathbf{0}_{\mathbf{p}}\right)$ to denote the model prediction score after replacing the masked-out context $\mathbf{x}_{-\mathbf{p}}$ with padding tokens $0_\mathbf{p}$:
\begin{equation}
    \label{eq:importance}
    \phi(\mathbf{p}, \mathbf{x}, e)=s(\mathbf{x},e)-s\left(\mathbf{x}_{-\mathbf{p}}, e ; \mathbf{0}_{\mathbf{p}}\right).
\end{equation}
However, the importance score $\phi(\mathbf{p}, \mathbf{x}, e)$ from equation~\ref{eq:importance} has a challenge that it is difficult to model the direct impact of $\mathbf{p}$ towards $e$ since the score is dependent on the context words around $\mathbf{p}$.
For example, in Figure.~\ref{fig:soc}, $\mathbf{p}$ (``\textit{the next mayor}'') is replaced by pad tokens to compute its importance towards the entity $e$ (``\textit{Cary Moon}'').
However, the importance score of $\mathbf{p}$ is dependent on context words around $\mathbf{p}$ (``\textit{won't be ... of Seattle}'').
To compute the context independent importance score, \texttt{SOC} proposes \textit{context sampling} that samples the context words around the phrase $p$ and computes the average prediction differences over the samples.
Specifically, it samples the context words $\hat{x}_{\delta}$ from a language model $p(\hat{x}_{\delta}|x_{-\delta})$ that is trained on $\mathcal{D}_{L}$, and obtains a set of context word replacements $\mathcal{S}$.
Here, we use 20 replacements.
For each replacement $\hat{x}_{\delta}\in\mathcal{S}$, we measure the prediction difference caused by replacing the phrase $\mathbf{p}$ with padding tokens. We take the average of these prediction differences to be the context-independent score $\phi(\mathbf{p}, \mathbf{x}, e)$ of the phrase $\mathbf{p}$, as expressed in equation \ref{eq:soc}:
\begin{equation} 
    \label{eq:soc}
        \scalebox{.85}{$
        \frac{1}{|\mathcal{S}|} \sum_{\hat{\mathbf{x}}_{\delta} \in \mathcal{S}}\left[s\left(\mathbf{x}_{-\delta},e ; \hat{\mathbf{x}}_{\delta}\right) - s\left(\mathbf{x}_{-\{\delta, \mathbf{p}\}},e ; \hat{\mathbf{x}}_{\delta} ; \mathbf{0}_{\mathbf{p}}\right)\right]$
        }
\end{equation}
In Figure.~\ref{fig:soc}, context words ``\textit{won't be}'' and ``\textit{of Seattle}'' around the phrase ``\textit{the next mayor}'' are replaced into ``\textit{will be}'' and ``\textit{of LA}'' which are sampled from the language model.
Then, the classifier computes the prediction difference between the sampled sentences with and without the phrase.

\paragraph{Phrase Selection}
The importance score $\phi(\mathbf{p}, \mathbf{x}, e)$ for phrase $\mathbf{p}$ is the degree to which $\mathbf{p}$ determines the correct entity type of target entity $e$.
After obtaining $\phi(\mathbf{p}, \mathbf{x}, e)$ for all phrase candidates $\mathcal{P}=\{\mathbf{p_i}\}$,
we pick the top $k$ candidate phrases with the highest importance score as the entity triggers, where $k$ is a hyperparameter.
Specifically, for each input instance $(\mathbf{x_i},\mathbf{y_i})\in\mathcal{D}_{L}$, we pick the top $k$ candidate phrases as entity triggers $T(\mathbf{x_i},\mathbf{y_i})$ per each entity $e\in\mathbf{x_i}$ to create a form $(\mathbf{x_i}, \mathbf{y_i}, T(\mathbf{x_i},\mathbf{y_i})) \in \mathcal{D}_T$.
Here, we search for the best $k$ out of [1, 2, 3, 5, 7, 10] (See Figure~\ref{fig:topk}).




\subsection{Trigger Interpolation Network (TIN)}
\label{ssec:tin}
The second stage of \autotrigger~ is the trigger interpolation network (\texttt{TIN}), which we define as a neural network that learns from a trigger-labeled dataset $\mathcal{D}_{T}$ consisting of a set of instances of the form $\{(\mathbf{x}, \mathbf{y}, T(\mathbf{x},\mathbf{y}))\}$.
The main idea behind the model is to combine two representations obtained from the sentence with target entity tokens and automatically extracted entity triggers.
In one representation, the entity tokens are masked and we hypothesize that it forces the model to infer the entity boundaries and types by leveraging the contextual clues.
However, some of the entity tokens itself may already have enough information to infer the entity type without any contextual clues.
Thus, in another representation, the entity tokens become the predominant source of information for classifying the entity as we mask out the triggers.

\begin{figure}[t!]
\vspace{-0.2cm}
    \centering 
    \includegraphics[scale=0.35]{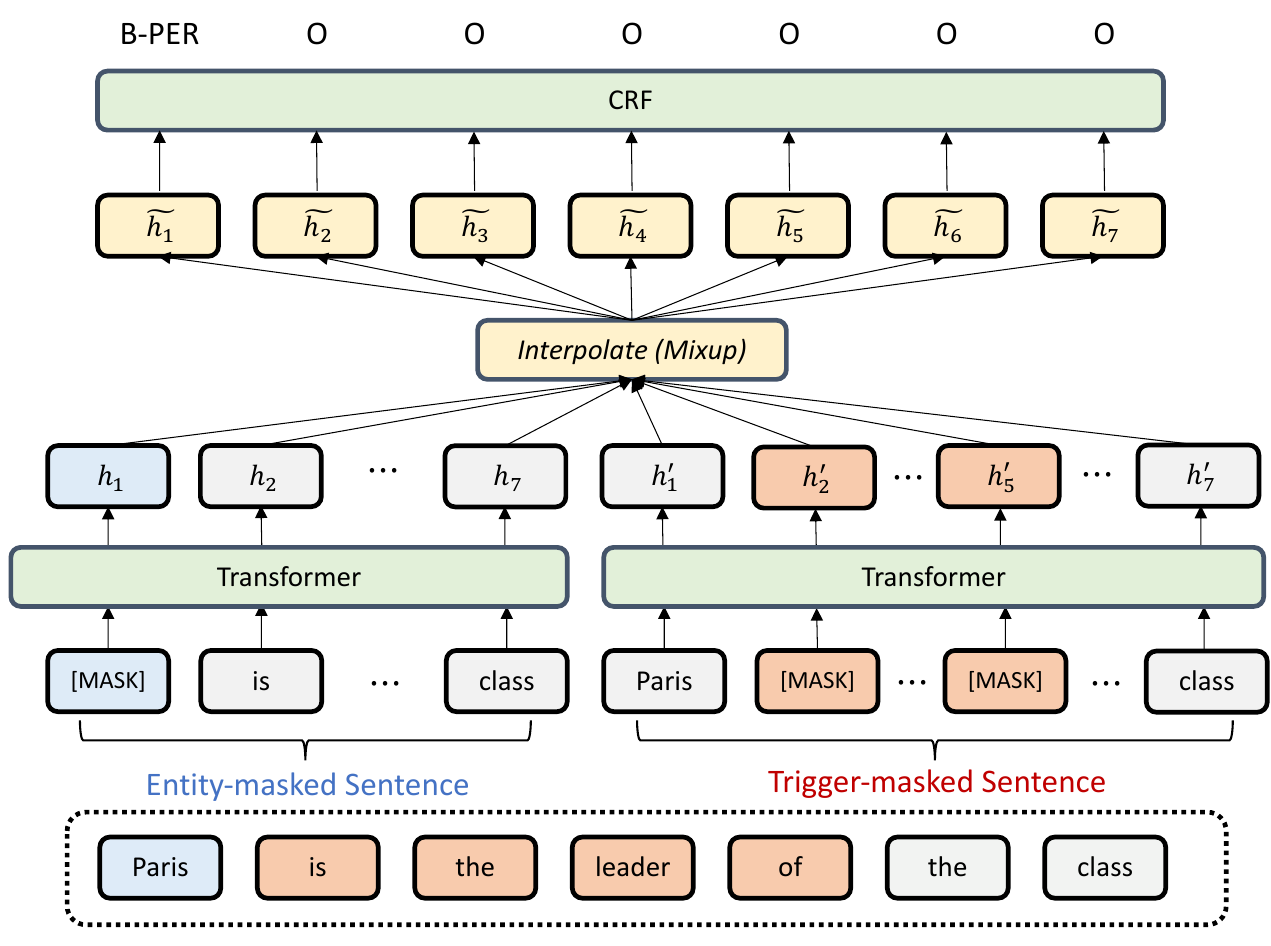}
    \vspace{-0.3cm}
    \caption{\textbf{Overview of the Trigger Interpolation Network (\texttt{TIN})}. Given an input sentence we create an Entity-masked sentence and a Trigger-masked Sentence. Then we interpolate token level representations $h_i$ and $h_i^\prime$ to create new hidden state representation, $\tilde{h_i}$. Interpolated hidden representations are fed to a CRF.
    }
    \label{fig:tin}
\end{figure}




\textsc{TIN} encodes the input sequence $\mathbf{x}$ with a transformer encoder $\mathbf{F}\left(.;\theta\right)$ and feeds the hidden representations $\mathbf{h}$ to a CRF tagger. 
Our proposal is to create two different representations of a token in a sequence and interpolate (mixup) them.
By doing this, we expect the model to predict less confidently on interpolations of hidden representations, which eventually make model be robust to both entity-perturbed and trigger-perturbed sentences.
As shown in Figure~\ref{fig:tin}, we first create entity-masked sentence $\mathbf{x}_{-e}$ and trigger-masked sentence $\mathbf{x}_{-t}$ for a given input instance $\{(\mathbf{x}, \mathbf{y}, T(\mathbf{x},\mathbf{y}))\}$, and then compute the interpolations in the output space of transformer encoder $\mathbf{F}\left(.;\theta\right)$ as follows:
\begin{equation}
\begin{aligned}
\mathbf{h}&=\mathbf{F}\left(\mathbf{x}_{-e} ; \theta\right),
\mathbf{h'}=\mathbf{F}\left(\mathbf{x}_{-t} ; \theta\right) \\
\tilde{\mathbf{h}}&=\lambda \mathbf{h}+(1-\lambda) \mathbf{h'}.
\end{aligned}
\end{equation}
Here, the transformer encoder $\mathbf{F}\left(.;\theta\right)$ for both $\mathbf{x}_{-e}$ and $\mathbf{x}_{-t}$ is sharing the weights.
Then we use interpolated hidden representations $\tilde{\mathbf{h}}$ as the input to the final CRF tagger.

When inferring tags on unlabeled sentences without entity triggers, 
we expect the trained $\mathbf{F}\left(.;\theta\right)$ to generate representation from the input $\mathbf{x}\in\mathcal{D}_u$, which can generalize well to unseen entities given the seen contextual clues, and seen entities given new surrounding context.
We then use it as an input to the final CRF tagger to get tag predictions.



\paragraph{Computational Cost.}
The computation overhead of training \texttt{TIN} is caused by forwarding input two times into the transformers to get two different representations, which is 1.5X larger than vanilla models (e.g. \texttt{BERT+TIN+CRF} vs. \texttt{BERT+CRF}).
However, the overhead of inference is same as vanilla models since the inference steps are same.

\section{Experimental Setup} 
\label{sec:exp}
In this section we describe datasets along with the baseline methods and experiment settings.

\vspace{-0.1cm}
\subsection{Datasets}
We consider three NER datasets as target tasks.
We consider two datasets for a bio-medical domain: \textbf{BC5CDR}~\citep{bc5cdr}, \textbf{JNLPBA}~\citep{kim2004introduction} and one dataset for a general domain: \textbf{CoNLL03}~\citep{conll}.
For BC5CDR and CoNLL03, we also have crowd-sourced entity trigger dataset $\mathcal{D}_{HT}$~\citep{lin-etal-2020-triggerner} to compare the quality of our automatically extracted triggers with.
They randomly sample 20\% of the data from each of the train sets and ask crowd-workers to select triggers for entities in those train sets.
Details on datasets are discussed in Appendix~\ref{ssec:data}.

\vspace{-0.1cm}
\subsection{Compared Methods}
To show the effectiveness of entity triggers, we compare baseline models that learn from entity-labeled dataset $\mathcal{D}_L$ and trigger-labeled dataset $\mathcal{D}_T$ respectively.
Here, we compare models under three different embedders: GloVE~\citep{Pennington2014GloveGV}, BERT~\citep{devlin-etal-2019-bert}, and RoBERTa~\citep{liu2019roberta}.
We don't include few-shot learning and semi-supervised learning due to following reasons:
(1) Few-shot learning~\cite{yang-katiyar-2020-simple} is a N-way k-shot task that needs N classes, each with k training examples.
This setup is restrictive because it requires a special sampling strategy since the number of training examples for the class must be fixed.
(2) Semi-supervised learning~\cite{Peng2019DistantlySN, autoner, yangner, Liu2019TowardsIN} uses additional supervisions and in-domain unlabeled data to augment train data by weakly labeling, which is different from our setup that only uses limited in-domain labeled data.

\vspace{-0.1cm}
\paragraph{Baselines without triggers}
We apply the following standard models on $\mathcal{D}_L$:
(1) \textbf{BLSTM+CRF} first adopts bidirectional LSTM to encode word embeddings into the representations.
Then, it feeds the representations into a linear layer to get label scores, and put the scores into a CRF tagger to predict the optimal path of entity tags;
(2) \textbf{CRF} considers word embeddings directly as the representations and conducts the same process of (1).


\paragraph{Methods with Triggers}
We apply the following models on $\mathcal{D}_T$:
(1) \textbf{BLSTM+TMN+CRF}~\citep{lin-etal-2020-triggerner} adopts the structured self-attention layer~\citep{selfattentive} above the bidirectional LSTM to encode the sentence and entity trigger into vector representation respectively.
It then learns trigger representations that can be generalized to unseen sentences to tag the named entity;
(2) \textbf{TMN+CRF} adopts the structured self-attention layer directly above the embeddings, and conducts the same process of (1);
(3) \textbf{TIN+CRF} is the \textit{trigger interpolation network} with CRF layer.


\begin{table*}[!t]
	\centering
	\small
	\resizebox{\textwidth}{!}{
		\begin{tabular}{llccccccccccccccc}
            \toprule
           \multirow{2}{*}{\textbf{Embedder}} &
           \multirow{2}{*}{\textbf{Method} / Train set \%} & \multicolumn{5}{c}{\textbf{BC5CDR}} & \multicolumn{5}{c}{\textbf{JNLPBA}} & \multicolumn{5}{c}{\textbf{CoNLL03}}\\
            \cmidrule(lr){3-7} \cmidrule(lr){8-12} \cmidrule(lr){13-17} & &
            20\% & 40\% & 60\% & 80\% & 100\% &
            20\% & 40\% & 60\% & 80\% & 100\% &
            20\% & 40\% & 60\% & 80\% & 100\% \\
            \midrule
            GloVE & + BLSTM+CRF & 71.92 & 76.29 & 79.04 & 80.72 & 81.07 & 66.36 & 69.31 & 71.25 & 71.90 & 72.79 & 85.06 & 88.33 & 88.98 & 89.84 & 90.72\\
            & + BLSTM+TMN+CRF & \bf 74.70 & \bf 78.15 & \bf 80.57 & \bf 82.77 & \bf 83.37 & \bf 66.78 & \bf 70.23 & \bf 71.41 & \bf 71.7 & \bf 72.55 & \bf 87.46 & \bf 88.88 & \bf 89.39 & \bf 90.16 & \bf 90.24 \\
            \midrule
            BERT & + BLSTM+CRF & 44.51 & 65.88 & 74.23 & 80.65 & 82.56 & 59.26 & 69.39 & 72.04 & 73.24 & 73.26 & 68.60 & 87.09 & 89.42 & 90.20 & 90.86\\
            & + CRF & 75.30 & 80.52 & 82.94 & 84.00 & 85.02 & 69.02 & 70.84 & 72.58 & 73.06 & 73.18 & \bf 88.61 & \bf 90.20 & \bf 91.10 & \bf 91.37 & \bf 91.48 \\
            & + BLSTM+TMN+CRF & 74.98 & 78.24 & 82.52 & 82.89 & 84.38 & 69.56 & 71.36 & 72.26 & 72.92 & 73.34 & 86.88 & 88.78 & 90.05 & 90.35 & 90.99 \\
            & + TMN+CRF & 75.07 & 80.14 & 82.57 & 84.03 & 85.49 & 69.25 & 71.34 & 71.70 & 72.98 & 73.38 & 88.39 & 89.34 & 90.24 & 90.83 & 91.06 \\
            & + TIN+CRF (Ours) & \bf 77.37 & \bf 81.40 & \bf 83.23 & \bf 85.25 & \bf 85.74 & \bf 70.48 & \bf 72.10 & \bf 72.81 & \bf 73.91 & \bf 74.83 & 87.84 & 89.64 & 89.71 & 90.39 & 90.75 \\
            \midrule
            RoBERTa & + CRF  & 82.85 & 84.63 & 86.08 & 86.44 & 87.10 & 71.07 & 72.19 & 73.32 & 73.50 & 75.37 & 90.53 & 91.63 & 91.90 & 92.06 & 93.09 \\
            & + BLSTM+TMN+CRF & 83.00 & 84.89 & 86.16 & 86.69 & 87.78 & 71.51  & 72.91 & 73.43 & 74.65 & 75.52 & 90.76 & 91.50 & 91.84 & 92.22 & 92.47 \\
            & + TMN+CRF & 83.89 & 85.56 & 86.65 & 87.31 & 87.78 & 71.79 & 72.87 & 73.10 & 74.19 & 75.13 & 90.82 & 91.47 & 92.32 & 92.42 & 92.57 \\
            & + TIN+CRF (Ours)  & \bf 84.45 & \bf 86.09 & \bf 87.5 & \bf 87.84 & \bf 88.09 & \bf 73.12 & \bf 74.23 & \bf 74.45 & \bf 74.96 & \bf 76.98 & \bf 91.37 & \bf 92.03 & \bf 92.63 & \bf 92.51 & \bf 93.24 \\
            \bottomrule
        \end{tabular}
	}
	\vspace{-0.2cm}
	\caption{\small Performance comparison (F1-score) by different percentage usage of the train data. For \texttt{TMN} and \texttt{TIN} baselines, we use the top 2 candidate phrases from SOC with constituency parsing as triggers. Best models under each embedder are \textbf{bold}.
	}
	\label{tab:overall}

\end{table*}

\begin{table*}[!t]
	\centering
	\small
	\resizebox{\textwidth}{!}{
		\begin{tabular}{llccccccccccccccc}
            \toprule
           \multirow{2}{*}{\textbf{Encoder}} &
           \multirow{2}{*}{\textbf{Method} / Train set \%} & \multicolumn{5}{c}{\textbf{BC5CDR}} & \multicolumn{5}{c}{\textbf{JNLPBA}} & \multicolumn{5}{c}{\textbf{CoNLL03}}\\
            \cmidrule(lr){3-7} \cmidrule(lr){8-12} \cmidrule(lr){13-17} & &
            20\% & 40\% & 60\% & 80\% & 100\% &
            20\% & 40\% & 60\% & 80\% & 100\% &
            20\% & 40\% & 60\% & 80\% & 100\% \\
            \midrule
            BERT & + CRF & 63.93 & 67.95 & 70.36 & 73.43 & 73.66 & 56.81 & 57.00 & 58.43 & 59.49 & 60.11 & 78.65 & 82.32 & 82.24 & 82.90 & 83.36 \\
            & + TMN+CRF & 65.37 & 68.51 & 68.80 & 72.17 & 73.46 & 56.39 & 56.98 & 58.14 & 59.01 & 59.98 & 80.18 & 82.39 & 82.16 & 82.46 & 83.93 \\
            & + TIN+CRF (Ours) & \bf 67.10 & \bf 69.21 & \bf 72.32 & \bf 73.81 & \bf 74.65 & \bf 57.19 & \bf 58.73 & \bf 59.67 & \bf 59.90 & \bf 60.36 & \bf 81.52 & \bf 82.82 & \bf 83.01 & \bf 83.72 & \bf 84.66  \\
            \midrule
            RoBERTa & + CRF & 72.16 & 74.55 & 75.92 & 76.39 & 76.17 & 58.88 & 60.05 & 61.22 & 62.32 & 63.71 & 82.39 & 84.06 & 84.50 & 84.98 & 85.61 \\
            & + TMN+CRF & 73.11 & 74.27 & 76.13 & 76.18 & 76.29 & 59.62 & 60.73 & 61.17 & 62.44 & 62.98 & 83.30 & \bf 85.30 & 85.71 & \bf 85.47 & 85.97 \\
            & + TIN+CRF (Ours) & \bf 73.65 & \bf 75.78 & \bf 77.14 & \bf 77.34 & \bf 76.90 & \bf 60.85 & \bf 61.50 & \bf 62.37 & \bf 63.55 & \bf 64.11 & \bf 84.11 & 84.82 & \bf 85.84 & 85.31 & \bf 87.38 \\
            \bottomrule
        \end{tabular}
	}
	\vspace{-0.2cm}
	\caption{\small Performance comparison (F1-score) on filtered test set by different percentage usage of the train data. Filtered test set is the set which the entities did not appear in the train and development sets. For \texttt{TMN} and \texttt{TIN} baselines, we use the top 2 candidate phrases from SOC with constituency parsing as triggers. Best models under each embedder are \textbf{bold}.
	}
	\label{tab:unseen}
	\vspace{-0.2cm}
\end{table*}
\section{Results and Analysis}
Here we look into a series of analysis questions as follows:
(1) Can \texttt{TIN} with automatically extracted triggers improve the performance of existing transformer-based models?
(2) Can \texttt{TIN} with automatically extracted triggers generalize well on the unseen entities?
(3) Are automatically extracted triggers work better than human-provided triggers?
(4) Can we improve the performance by asking human to refine automatically extracted triggers?
We present experiment settings and details in Appendix~\ref{ssec:setting}-~\ref{ssec:metric}.

\subsection{Performance Comparison}
\label{sssec:overall}
\vspace{-0.0cm}
Here, we test all models by varying the amount of training data from 20\% to 100\% to show the impact of train data size. 
In Table~\ref{tab:overall}, we report the performance of the baseline approaches and our model variants on three different datasets.
Here are findings:
(1) \texttt{BLSTM} harms the performance of transformer-based models while it works well with GloVE embeddings;
(2) Models that receive both entities and triggers as input generally outperform the \textit{entity-only} baselines (See \texttt{CRF} vs. \texttt{TIN+CRF}, \texttt{TMN+CRF});
(3) \texttt{TIN} outperforms \texttt{TMN}, which shows that \texttt{TIN} is doing better leveraging entity triggers for NER, and \texttt{RoBERTa+TIN+CRF} outperforms all the baselines regardless of the amount of data that is used to train it;
(4) Comparing \texttt{BERT+TIN+CRF} to \texttt{BERT+CRF}, we observe a performance drop in CoNLL03. 
We further investigate this phenomenon and find a large drop in F1 score (from 0.82 to 0.79) for the MISC class from the \texttt{BERT+TIN+CRF} (See Table~\ref{tab:classificationreport}), which shows that triggers may provide a precision decreasing signal for the MISC type.

\begin{table}[t]
\vspace{0.1cm}
	\centering
	\scalebox{0.68 
	}{
		\begin{tabular}{ccccccc}
			\toprule
			\textbf{Type} & \multicolumn{3}{c}{\texttt{BERT+CRF}} & \multicolumn{3}{c}{\texttt{BERT+TIN+CRF}} \\
			\cmidrule(lr){2-4} \cmidrule(lr){5-7}
			& Precision & Recall & F1-score & Precision & Recall & F1-score \\
			\midrule
			LOC & 0.92 & 0.94 & 0.93 & 0.91 & 0.93 & 0.92 \\
			MISC & 0.81 & 0.82 & \bf 0.82 & 0.75 & 0.84 & \underline{0.79} \\
			ORG & 0.88 & 0.90 & 0.89 & 0.86 & 0.90 & 0.88 \\
			PER & 0.97 & 0.96 & 0.96 & 0.96 & 0.96 & 0.96 \\
			\bottomrule
		\end{tabular}
	} 
	\caption{Classification Report (F1-score) of \texttt{BERT+CRF} and \texttt{BERT+TIN+CRF} on CoNLL03.}
	\label{tab:classificationreport}
	\vspace{-0.4cm}
\end{table}

\paragraph{Performance in extreme label scarcity settings}
\label{sssec:low}

\begin{figure}[t]
\hspace{0.1cm}
\begin{minipage}{0.23\textwidth}
  \centerline{\includegraphics[width=4.1cm]{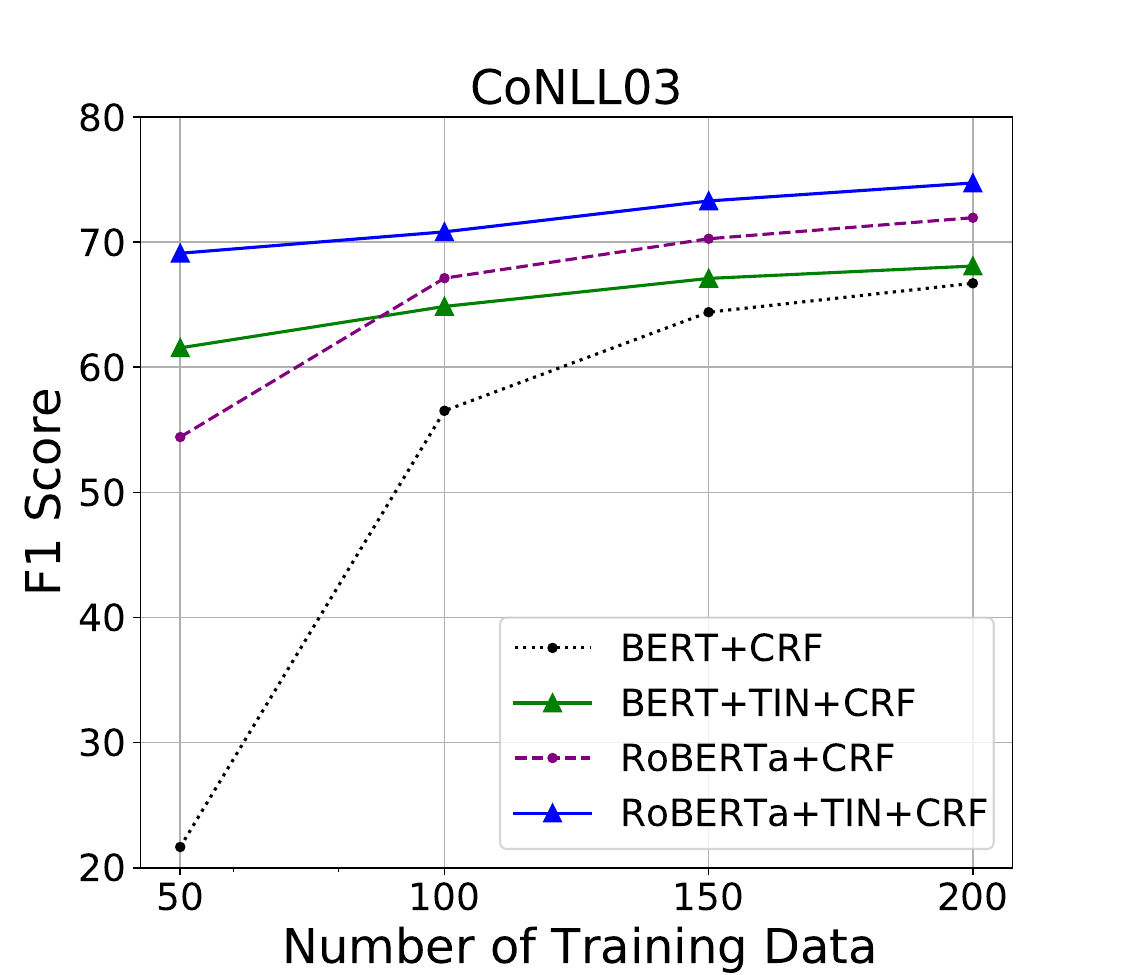}}
  \vspace{-0.1cm}
  \centerline{\small (a) CoNLL03}
\end{minipage}
\begin{minipage}{0.23\textwidth}
  \centerline{\includegraphics[width=4.1cm]{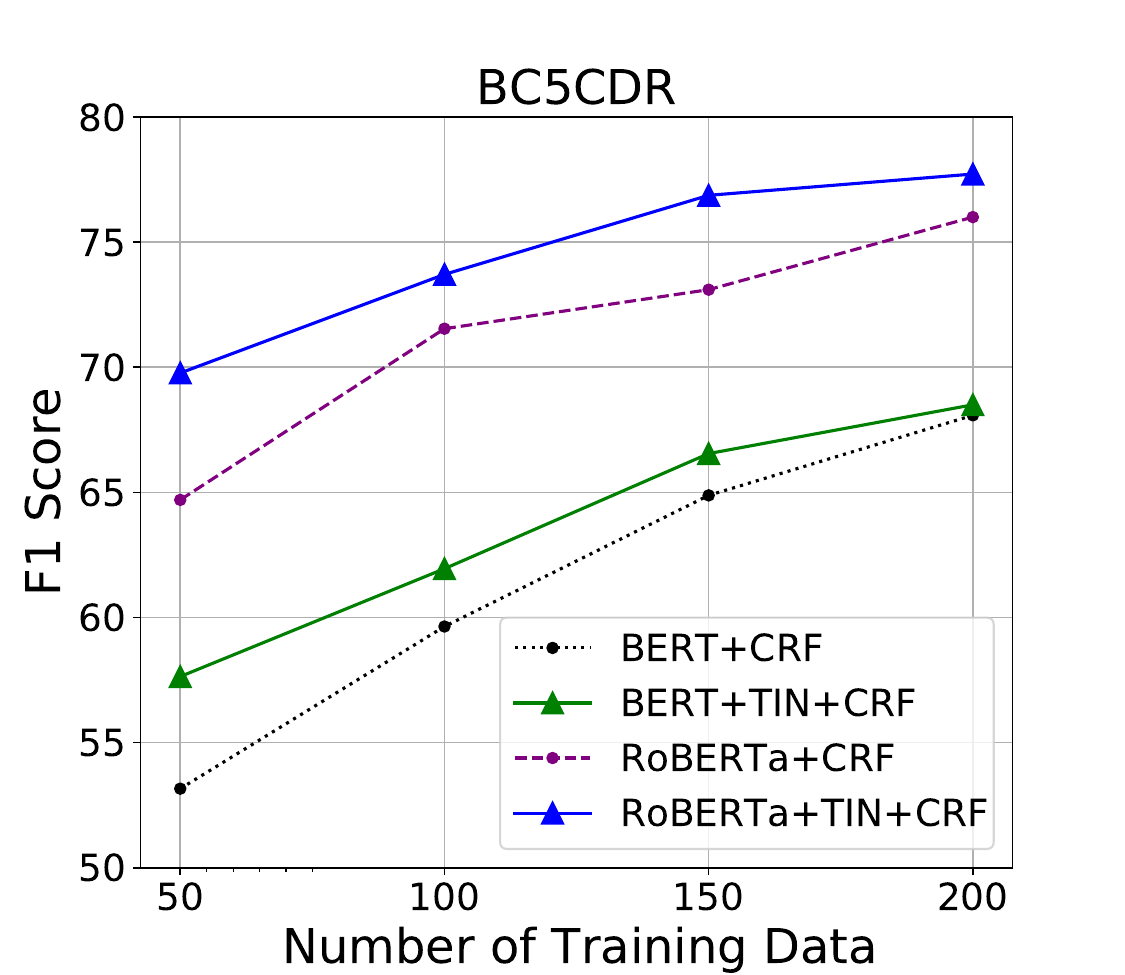}}
  \vspace{-0.1cm}
  \centerline{\small (b) BC5CDR}
\end{minipage}
  \vspace{-0.2cm}
\caption{Performance Comparison (F1-score) on CoNLL03 and BC5CDR by small numbers of train data instances (50, 100, 150, 200).
}
\label{fig:low}
\vspace{-0.2cm}
\end{figure}

We hypothesize that our models will have larger performance gains in extreme label scarcity settings, because of their ability to leverage additional information from triggers which enables them to reap more benefits from given training data. 
To investigate this we observe the performance of our models and baselines starting with only 50-200 sentences to train them.
Figure~\ref{fig:low} shows the performance under the extreme label scarcity setting. 
Here, \texttt{TIN+CRF} achieves large performance gain in extremely low-resource setting. 
Specifically, we observe over 50\% relative gain compared to the baseline for 50 training sentences.

\paragraph{Unseen entity generalization}
We hypothesize that our framework can generalize well to unseen entities not presented in the training data, attribute to prior knowledge of contextual clues.
To evaluate the generalizability of the model, we create a filtered test set which contains instances with only entities that do not appear in the train data.
Table~\ref{fig:low} shows the performance of model variants on the filtered test set by different amount of training data. 
Here, \texttt{TIN+CRF} shows its effectiveness on predicting unseen entities, and much more when we compare under the small amount of training data.

\subsection{Human-in-the-loop Trigger Extraction}
\label{sssec:humanauto}
We aim to study whether human participation in trigger creation could be helpful.
First, we compare the performance of model variants trained with automatically extracted triggers (\texttt{auto}) and human-provided triggers (\texttt{human}), and check its label efficiency.
Then, we conduct a small-scale experiment of trigger refinement by human annotators.

\begin{figure}[t]
\centerline{\includegraphics[width=7.5cm]{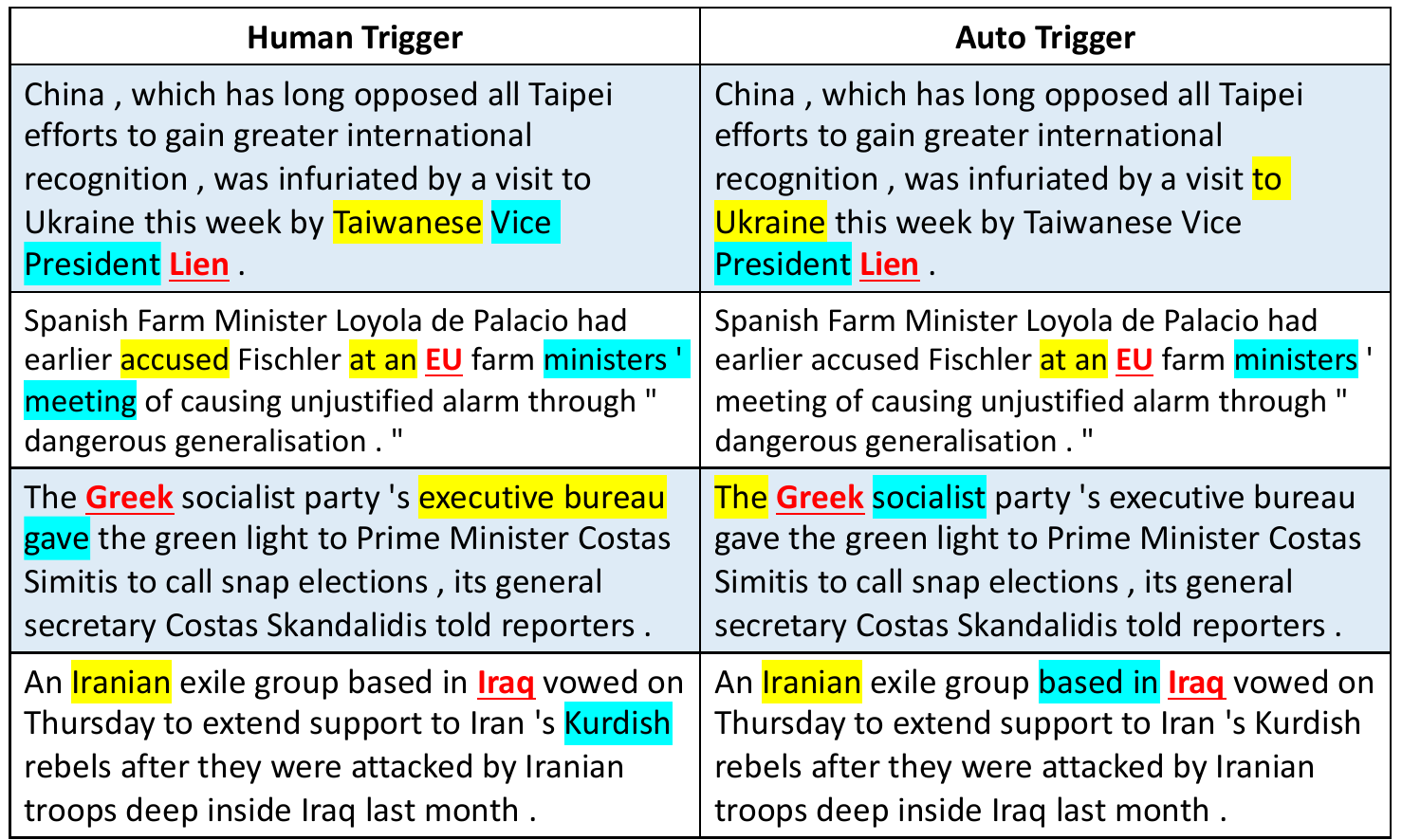}}
\vspace{-0.2cm}
\caption{Top 2 highlighted \texttt{auto} and \texttt{human} triggers corresponding to the boldfaced and underlined entity.}
\label{fig:triggercase}
\vspace{-0.2cm}
\end{figure}

\smallskip
\noindent
\textbf{Human-curated vs. Auto Triggers.}
We use $\mathcal{D}_{HT}$ as the source of human triggers and use the same dataset to extract auto triggers with \texttt{SOC} algorithm (see Appendix Table~\ref{tab:dataset}).
We then sample 25\%, 50\%, and 75\% of the instances from both to construct 5\%, 10\%, 15\% percent of our experimentation dataset (since $\mathcal{D}_{HT}$ is a 20\% random sample from $\mathcal{D}_{L}$).
One big difference between \texttt{human} and \texttt{auto} is whether the triggers are contiguous token spans or not.
For example, humans are asked to annotate a group of word tokens that represent \textit{``general''} phrase like \textit{``had dinner at''} from the sentence \textit{``We \underline{had} a fantastic \underline{dinner at} Sunnongdan.''}, while a set of phrase candidates $\mathcal{P}$ from the constituency parse tree can only contain the contiguous token spans.
Figure.~\ref{fig:triggercase} shows examples of \texttt{human} and \texttt{auto}.
These examples are from CoNLL03, and \texttt{auto} are extracted from the entity token classifier which is trained on 20\% of the train data.
Tab.~\ref{tab:autohuman} shows that \texttt{auto} are comparable or even stronger than \texttt{human} even though created with no human labeling.
The success of auto triggers might be that their impact on the entity labeling is directly at the model level, while human triggers, even if they are meaningful on the surface level, might have less impact in determining the entity label as they do not mimic what the model thinks. 
We manually inspect the \texttt{auto} and \texttt{human} and found that \texttt{auto} are consecutive while \texttt{human} are usually non-consecutive. 
Even though there could be many reasons for the sub-optimal performance of human selected triggers available in the dataset  \citep{lin-etal-2020-triggerner}, we do not rule out the possibility of leveraging human expertise to help.
Further case examples are presented in Appendix~\ref{ssec:ablation}.

\begin{table}[t]
	\centering
	\scalebox{0.55 
	}{
		\begin{tabular}{ccccccc}
			\toprule
			\textbf{BC5CDR} & \multicolumn{2}{c}{\texttt{BLSTM+TMN+CRF}} & \multicolumn{2}{c}{\texttt{BERT+TIN+CRF}} & \multicolumn{2}{c}{\texttt{RoBERTa+TIN+CRF}}\\
			\cmidrule(lr){2-3} \cmidrule(lr){4-5} \cmidrule(lr){6-7} Percentage / \textbf{Model} & \texttt{human} & \texttt{auto} & \texttt{human} & \texttt{auto} & \texttt{human} & \texttt{auto} \\
			\midrule
			5\% & \bf 26.96 & 24.70 & 66.20 & \bf 66.50 & 75.79 & \bf 76.92 \\
			10\% & \bf 46.24 & 43.54 & 71.25 & \bf 71.84 & 80.92 & \bf 81.63 \\
			15\% & \bf 51.29 & 50.44 & 73.88 & \bf 74.11 & 83.54 & \bf 83.87 \\
			20\% & \bf 56.28 & 54.91 & 75.97 & \bf 76.58 & 83.88 & \bf 84.17 \\
			\bottomrule
		\end{tabular}
	} 
	\scalebox{0.55 
	}{
		\begin{tabular}{ccccccc}
			\toprule
			\textbf{CoNLL03} & \multicolumn{2}{c}{\texttt{BLSTM+TMN+CRF}} & \multicolumn{2}{c}{\texttt{BERT+TIN+CRF}} & \multicolumn{2}{c}{\texttt{RoBERTa+TIN+CRF}}\\
			\cmidrule(lr){2-3} \cmidrule(lr){4-5} \cmidrule(lr){6-7} Percentage / \textbf{Model} & \texttt{human} & \texttt{auto} & \texttt{human} & \texttt{auto} & \texttt{human} & \texttt{auto} \\
			\midrule
			5\% & 56.39 & \bf 57.95  & 78.17 & \bf 78.56 & 84.72 & \bf 85.71\\
			10\%  & 61.89 & \bf 66.58 & 81.67 & \bf 82.19 & 87.80 & \bf 88.12 \\
			15\% & 67.48 & \bf 69.41 & 83.67 & \bf 85.13 & 88.40 & \bf 89.68 \\
			20\% & 71.11 & \bf 74.43 & 84.88 & \bf 85.58 & 89.68 & \bf 90.21 \\
			\bottomrule
		\end{tabular}
	} 
    \vspace{-0.2cm}
	\caption{Performance comparison (F1-score) of \texttt{TMN} and \texttt{TIN} with \texttt{human} and \texttt{auto} triggers.}
	\label{tab:autohuman}
    \vspace{-0.5cm}
\end{table}

\begin{figure}[t]
\centerline{\includegraphics[width=8.8cm]{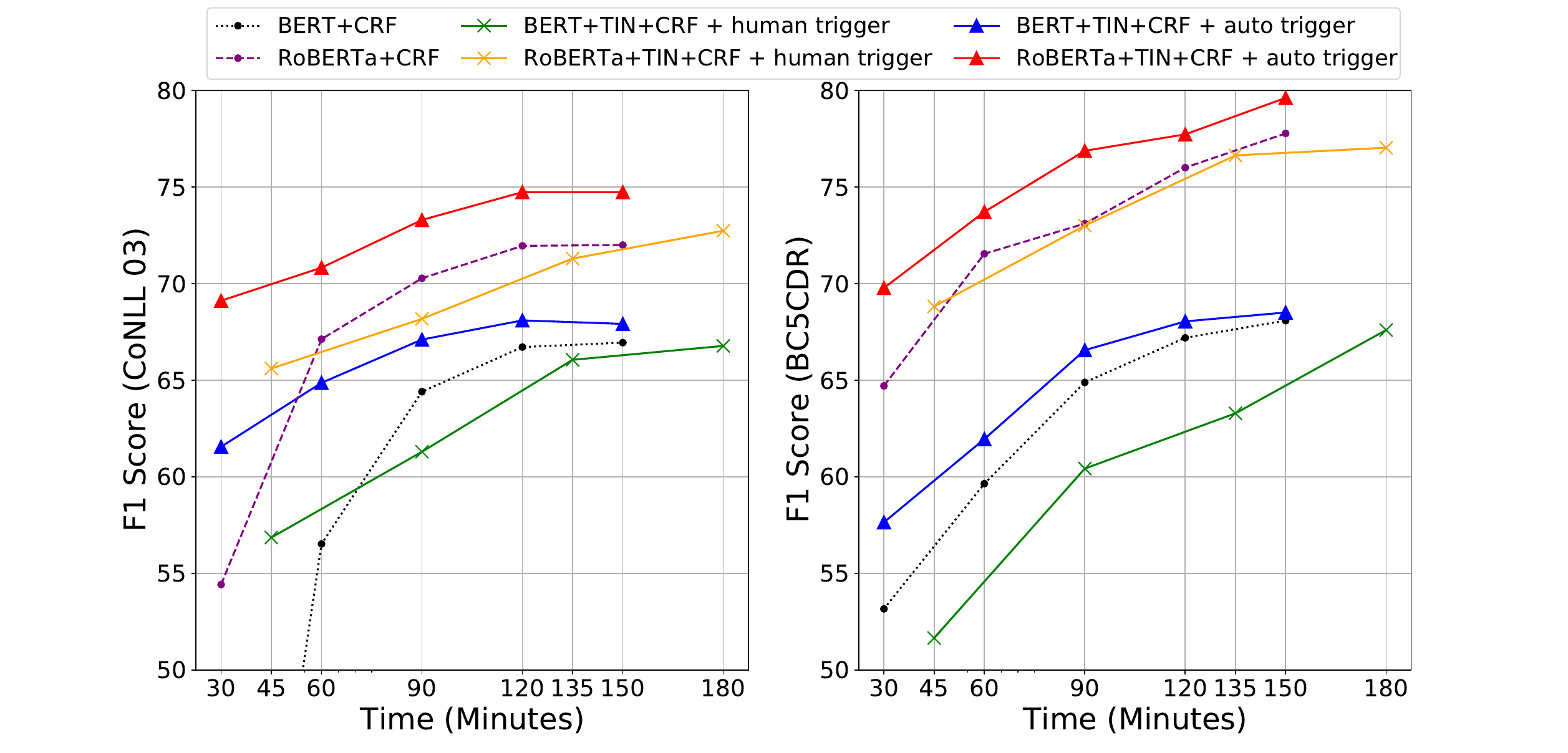}}
\vspace{-0.3cm}
\caption{Performance Comparison (F1-score) by annotators' labeling time cost.}
\label{fig:labeleff}
\vspace{-0.4cm}
\end{figure}

\paragraph{Efficiency of Human-curated vs. Auto Triggers.}
\label{sssec:labeleff}
We conduct a study to measure how trigger extraction by human affects labeling efficiency.
First, we found that labeling triggers along with the entities is 1.5 times slower than labeling the entities only.
Given this observation, we compare the performance between TIN models with 
\texttt{human} and \texttt{auto} by holding annotation time constant.
We present the study in Figure.~\ref{fig:labeleff}.
Each marker on the x-axis of the plots indicate a certain annotation time, which is represented by approximate time.
First, we could find that if we ask human to get triggers from scratch, it may not work better (See \texttt{RoBERTa+CRF} vs. \texttt{RoBERTa+TIN+CRF+human}).
However, if we use auto triggers, we could achieve better performance and label efficiency (See \texttt{RoBERTa+TIN+CRF+auto}).

\begin{figure}[t]
\begin{minipage}{0.23\textwidth}
  \centerline{\includegraphics[width=4.1cm]{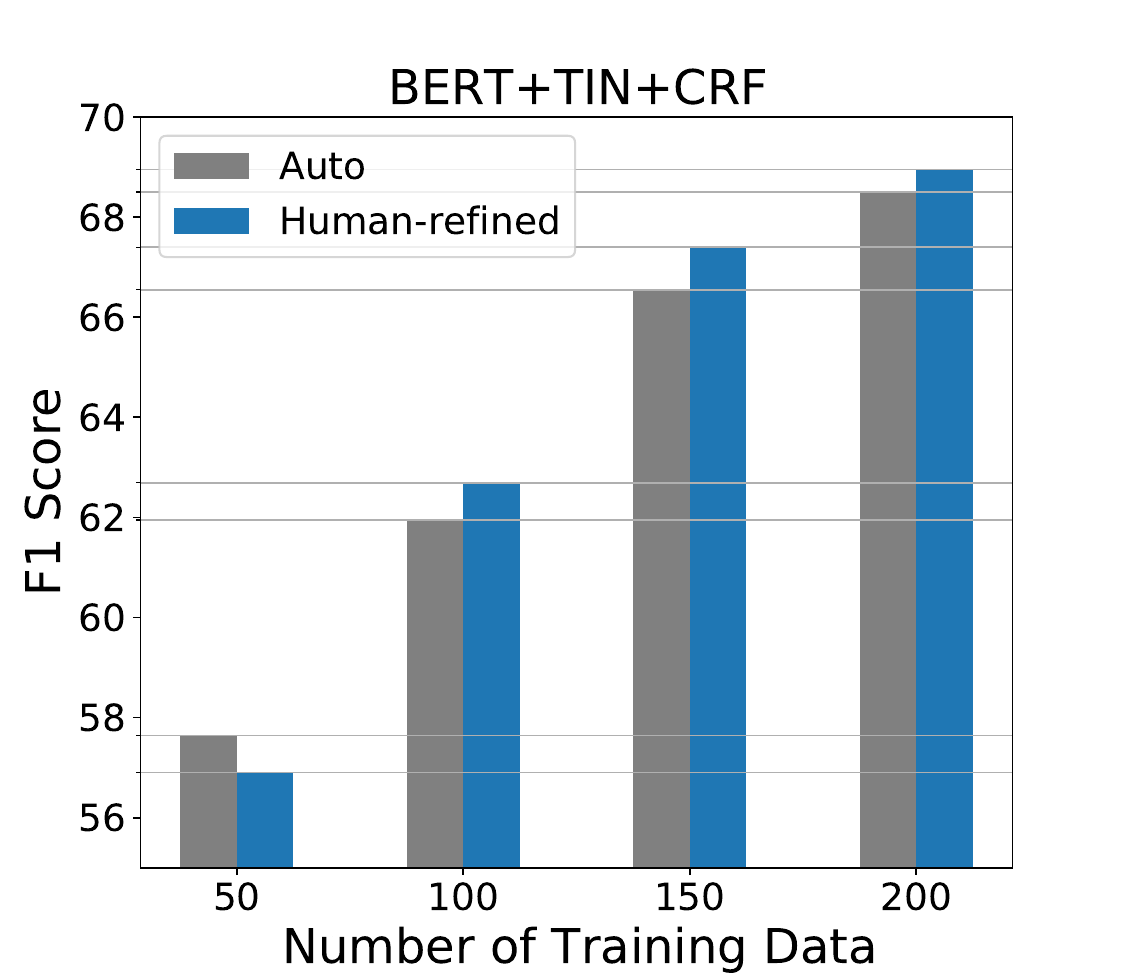}}
\end{minipage}
\begin{minipage}{0.23\textwidth}
  \centerline{\includegraphics[width=4.1cm]{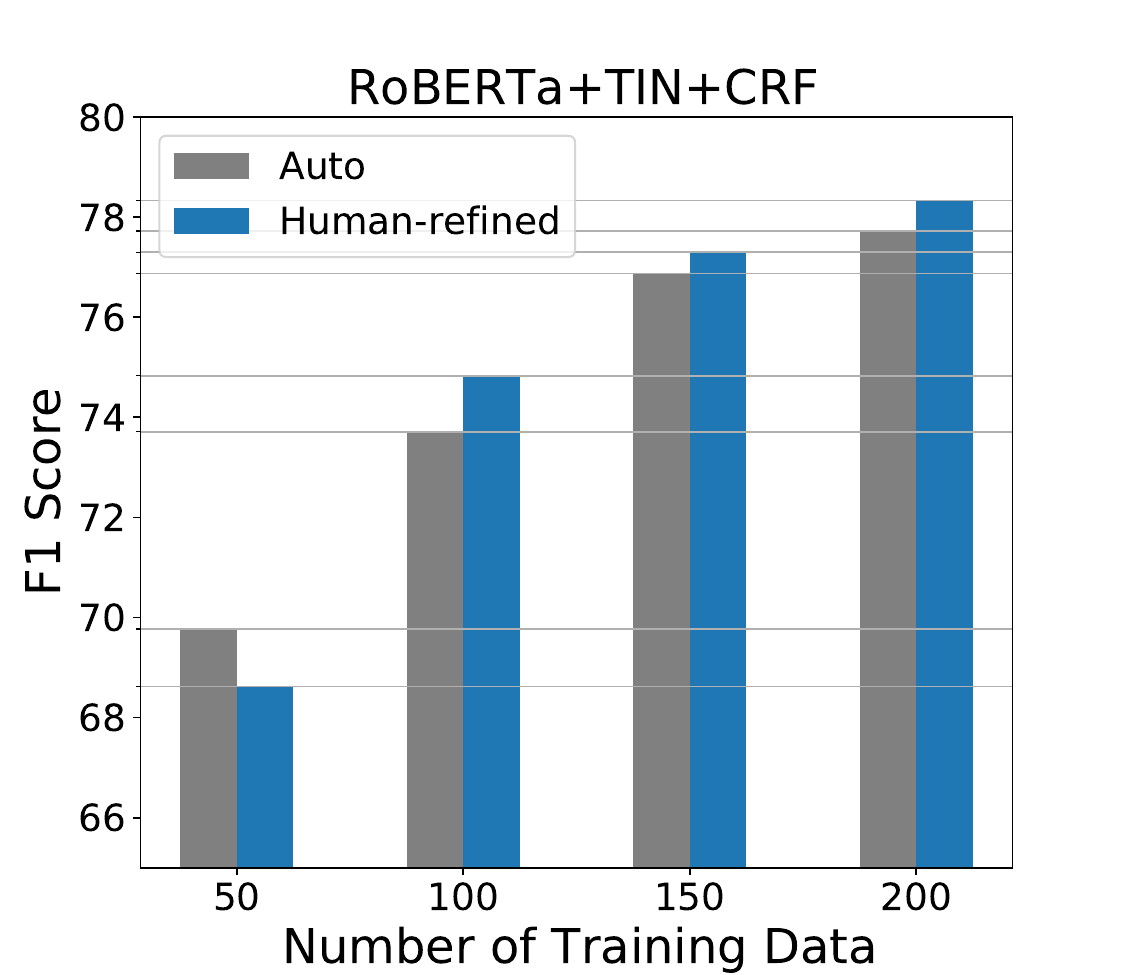}}
\end{minipage}
\vspace{-0.2cm}
\caption{Performance Comparison (F1-score) on BC5CDR by different numbers of train data (50, 100, 150, 200) with auto and human-refined auto triggers.
}
\label{fig:hitl}
\vspace{-0.5cm}
\end{figure}

\smallskip
\noindent
\textbf{Human-in-the-loop Trigger Refinement.}
In previous study, we could find that asking human to annotate triggers from the scratch is not efficient.
Here, we further think about asking human to refine auto triggers to reflect human decisions in a less labor intensive way.
For all our previous experiments, we use the top two auto triggers, which limits our capacity to make the best use of them.
In this experiment, given a training set with labeled entities, we extract five auto triggers (Sec.~\ref{ssec:soc}), show them to a human in a minimal interface, and ask for relevance judgments (relevant/non-relevant). 
We judge relevance of the automatically extracted triggers for entities in 50 - 200 sentences.
Figure.~\ref{fig:hitl} shows that we get an additional performance boost with more than 50 training sentences, when human-refined auto triggers are used.
It shows promise for blending human expertise with auto triggers. 




\section{Conclusion} 
\label{sec:conclusion}
In this paper, we proposed a novel two-stage framework to generate and leverage explanations for named entity recognition.
It automatically extracts essentially human-readable clues in the text, which is called entity triggers, by sampling and occlusion algorithm and leverages these triggers with trigger interpolation network.
The model effective learns the prior knowledge to infer the entity boundaries and types by leveraging the contextual clues.
We showed that our framework, named \textsc{AutoTriggER}, successfully generates entity triggers and effectively leverages them to improve the overall performance, especially in the label scarcity setting for technical domains where domain-expert annotations are very limited due to the high cost.
Moreover, it shows better generalizability on unseen entities that do not appear in training data.
We believe that this work opens up future works that can be extended to semi-supervised learning or distant supervised learning which can effectively use automatically extracted triggers to weakly label the unlabeled corpus.
\section*{Limitations}
\textsc{AutoTriggER} could reduce human efforts on collecting rationales for NER to improve the label efficiency and unseen entity generalizability of the model.
However, it has a limitation that the time cost of extracting entity triggers is pretty expensive (i.e., (1) train entity token classifier; (2) run constituency parsing to retrieve trigger candidates; and (3) run post-hoc explanation to assign importance score to each candidate).
This limitation raises open questions about whether reducing total training time cost or human effort is more efficient and important.

\section*{Acknowledgement}
This work was funded by the Defense Advanced Research Projects Agency with award W911NF-19-20271.



\bibliography{anthology,custom,sample-base}
\bibliographystyle{acl_natbib}

\clearpage
\appendix
\section{Appendix} 
\label{sec:appendix}

\subsection{Experiment Settings}
\label{ssec:setting}
We implement all the baselines using PyTorch~\citep{NEURIPS2019_9015} and HuggingFace~\citep{wolf-etal-2020-transformers}.
We set the batch size and learning rate to 10 and 0.01 for GloVE embedder models (i.e., \texttt{GloVE+BLSTM+CRF}, \texttt{GloVE+TMN+CRF}) while we set 30 and 2e-5 for all other transformer embedder models (See Table \ref{tab:setting}).
Note that for experiments in extreme low resource setting (Sec. \ref{sssec:low}), we set the batch size to 4 for training the models due to the extremely limited training data. 
\paragraph{Trigger Matching Network}
For transformer-based \texttt{TMN} (i.e., \texttt{BERT+BLSTM+TMN+CRF}, \texttt{BERT+TMN+CRF}, etc.) we re-implement since the original repository does not support transformer embedders.
\paragraph{Trigger Interpolation Network}
For our \texttt{TIN}, we set the interpolation $\lambda$ to 0.5.
\paragraph{Automatic Trigger Extraction}
For automatic trigger extraction stage, we build the entity token classifier with cased BERT-base encoder for \texttt{BERT+TIN+CRF} and RoBERTa-large for \texttt{RoBERTa+TIN+CRF}.
The entity token classifier consists of the transformer encoder to encode each word token followed by a token-level linear layer that classifies each token to an entity tag.
We use a batch size of 16 and learning rate of 1e-4 for training.
For experiments under extreme low resource setting, we set batch size to 4 similar to the \texttt{TIN} models.
To run context sampling in the SOC algorithm, we use a LSTM language model which is pre-trained on the training data.


\subsection{Evaluation Metrics}
\label{ssec:metric}
We evaluate our framework by recall (R), precision (P), and F1-score (F1), though only report F1 in these experiments.
Recall (R) is the number of correctly recognized named entities divided by the total number of named entities in the corpus,
and precision (P) is the number of correctly recognized named entities divided by the total number of named entities recognized by the framework.
A recognized entity is correct if both its boundary and its entity type are exact matches to the annotations in the test data.
F1-score is the harmonic mean of precision and recall.

\begin{table}[t]
	\centering
	\small
	\scalebox{0.9}{
		\begin{tabular}{cccc}
			\toprule   
			\multirow{2}{*}{\textbf{Embedder}} &      
			\multirow{2}{*}{GloVE}  &  
			\multicolumn{2}{c}{Transformer} \\
			\cmidrule{3-4} & & BERT & RoBERTa \\
            \midrule
            batch size & 10 & 30 & 30 \\
            learning rate & 0.01 & 2e-5 & 2e-5 \\
            epochs & 10 & 10 & 10 \\
            LSTM hidden dimension & 200 & - & - \\
			\bottomrule
		\end{tabular}
	} 
	\caption{Experimental setting details.}
	\label{tab:setting}
\end{table}

\subsection{Data Statistics}
\label{ssec:data}
\textbf{BC5CDR}~\citep{bc5cdr} is a bio-medical domain NER dataset from BioCreative V Chemical and Disease Mention Recognition task. It has 1,500 articles containing 15,935 \textsc{Chemical} and 12,852 \textsc{Disease} mentions.
\textbf{JNLPBA}~\citep{kim2004introduction} is a bio-medical domain NER dataset for the Joint Workshop on NLP in Biomedicine and its Application Shared task. It is widely used for evaluating multiclass biomedical entity taggers and it has 14.6K sentences containing \textsc{protein}, \textsc{DNA}, \textsc{RNA}, \textsc{cell line} and \textsc{cell type}.
\textbf{CoNLL03}~\citep{conll} is a general domain NER dataset that has 22K sentences containing four types of general named entities: \textsc{location}, \textsc{person}, \textsc{organization}, and \textsc{miscellaneous} entities that do not belong in any of the three categories.

\subsection{Performance Analysis}
\label{ssec:ablation}
\begin{figure}[t]
\hspace{0.1cm}
\begin{minipage}{0.22\textwidth}
  \centerline{\includegraphics[width=4.1cm]{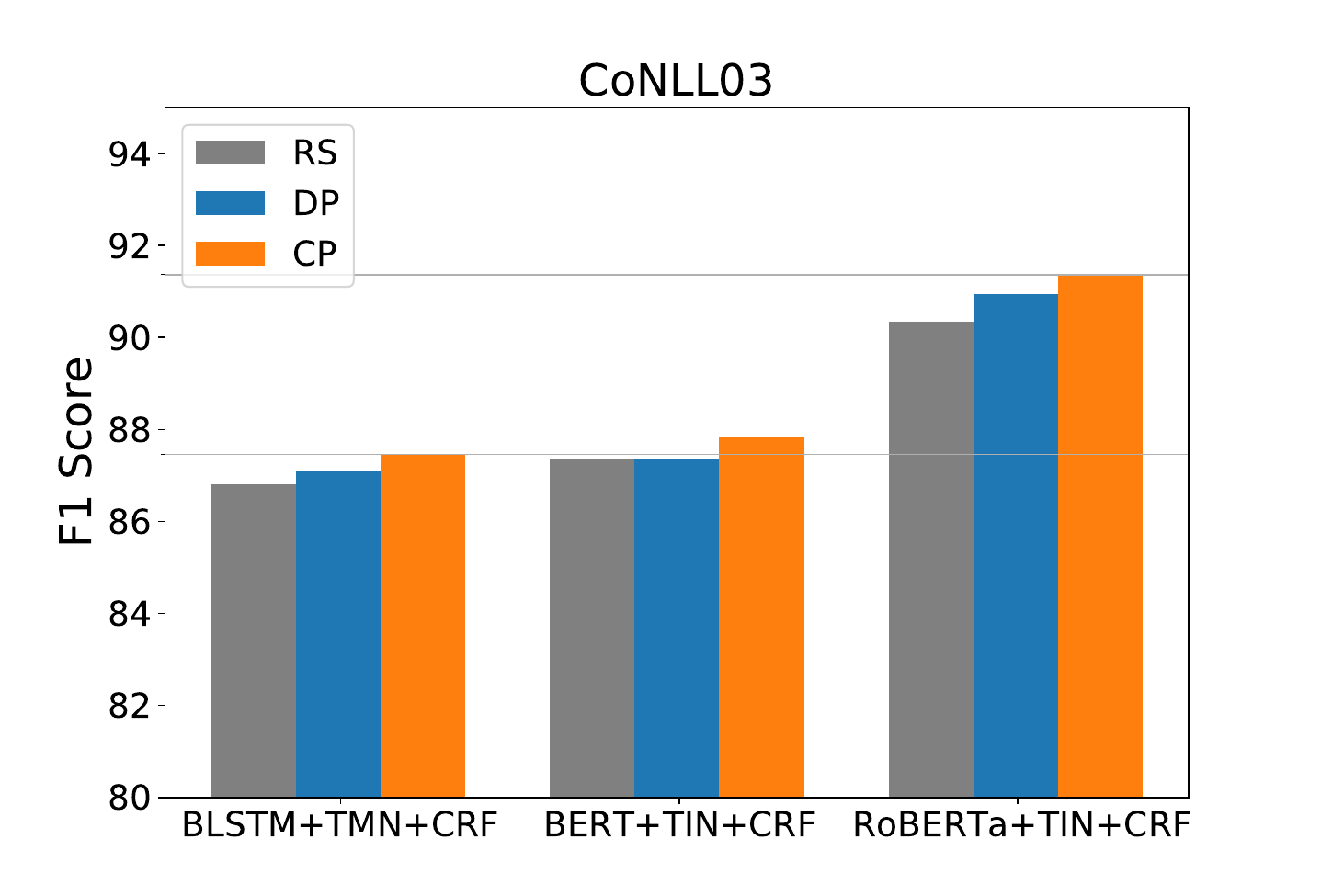}}
  \vspace{-0.1cm}
  \centerline{\small (a) CoNLL03}
\end{minipage}
\hspace{0.1cm}
\begin{minipage}{0.22\textwidth}
  \centerline{\includegraphics[width=4.3cm]{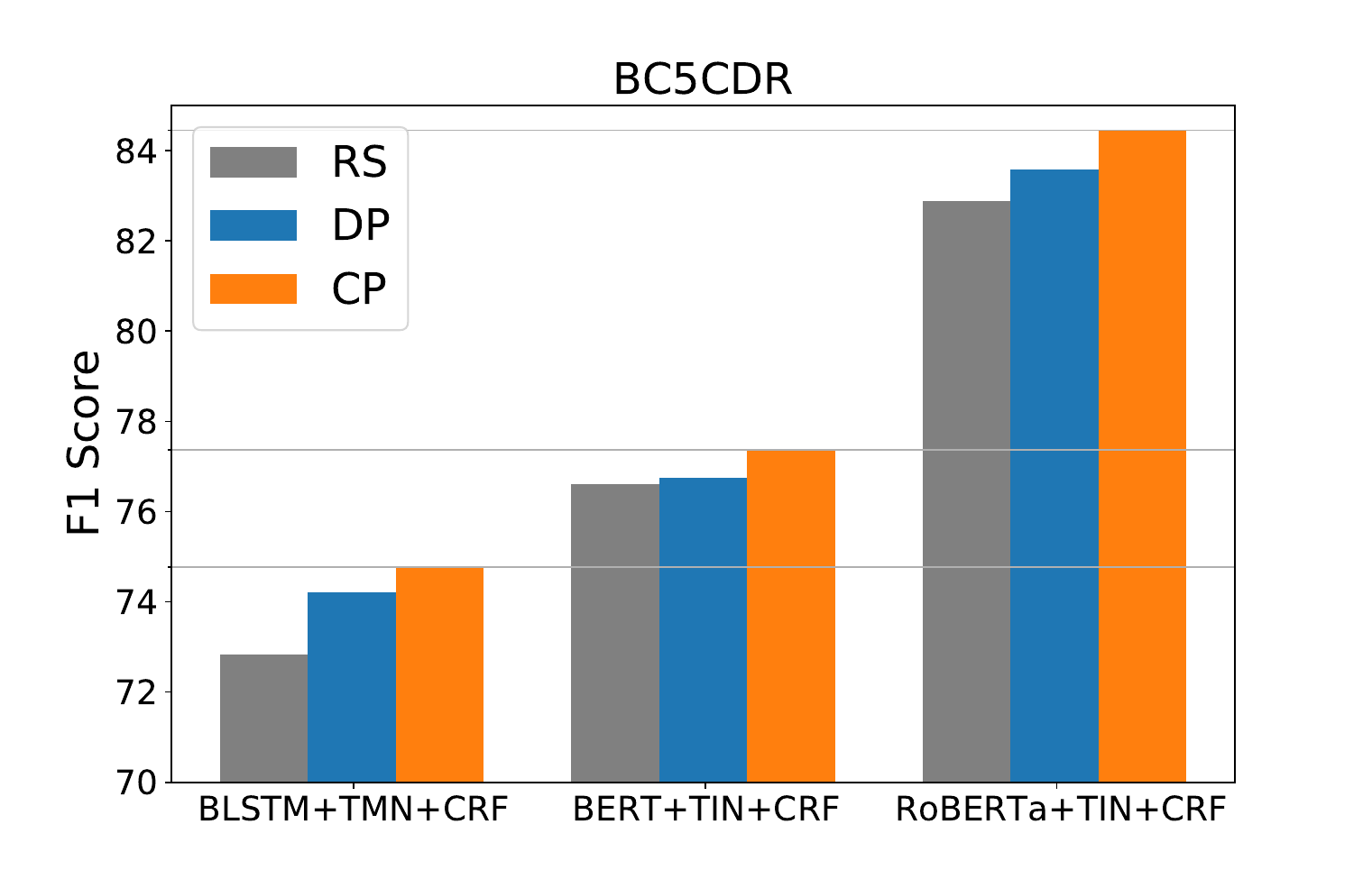}}
  \vspace{-0.1cm}
  \centerline{\small (b) BC5CDR}
\end{minipage}
\caption{Performance comparison (F1-score) of entity+trigger baselines on 20\% training dataset of CoNLL03 and BC5CDR with different trigger candidate variants.}
\label{fig:triggervariants}
\vspace{-0.2cm}
\end{figure}

\smallskip
\noindent
\textbf{Trigger Candidate Variants.}
In Sec~\ref{ssec:soc}, we first constructed a set of phrase candidates $\mathcal{P}$ for which the importance score is computed.
To show the efficacy of constituency parsing for constructing trigger candidates, we conduct an ablation study on different variants of it.
For the construction, we compare three variants:
(1) \texttt{RS} is random selection. It randomly chooses $n$ contiguous tokens to be grouped as a phrase for $k$ times.
Consequently, $\mathcal{P}$ is composed of $k$ random spans.
(2) \texttt{DP} is dependency parsing. Here, to generate $\mathcal{P}$, we first parse the input sentence using dependency parsing. Then, we traverse from the position of entity mention in the input sentence using depth-first-traversal and get a list of tokens visited for each hop up to 2-hops. Finally, for each hop, we convert the list of tokens to a list of phrases by merging the tokens that are contiguous into a single phrase. 
(3) \texttt{CP} is constituency parsing, which is our current method (see Sec.~\ref{ssec:soc}).
We expect each variant to provide different syntactic signals to our framework.
Figure~\ref{fig:triggervariants} shows the model's performance with triggers that have been selected from different sets of phrase candidates.
As we can see, constituency parsing yields consistently better performance by providing better quality of syntactic signals than others.

\begin{figure}[t!]
\hspace{0.1cm}
\begin{minipage}{0.23\textwidth}
  \centerline{\includegraphics[width=4.2cm]{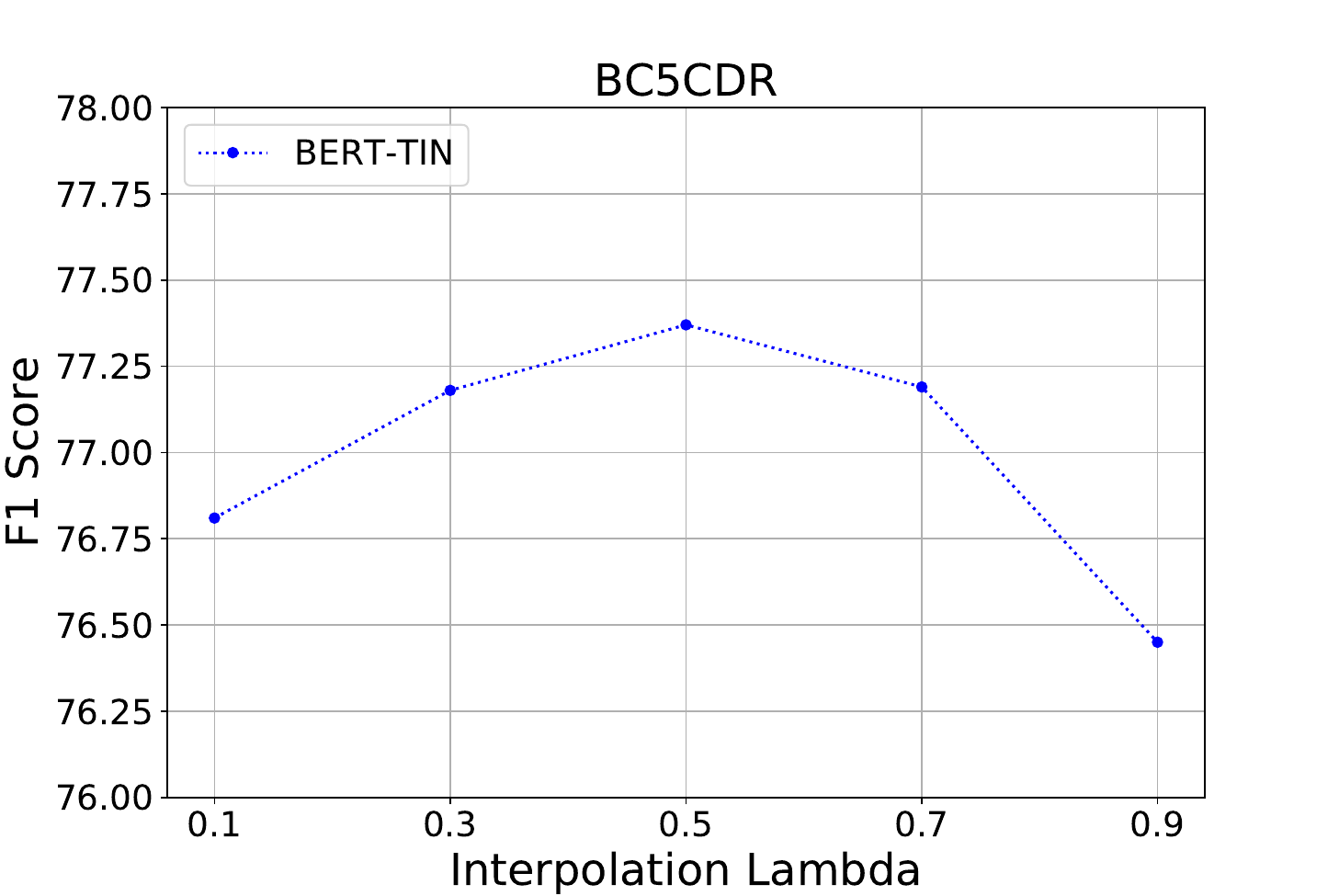}}
  \vspace{-0.1cm}
  \centerline{\small (a) BERT+TIN+CRF}
\end{minipage}
\begin{minipage}{0.23\textwidth}
  \centerline{\includegraphics[width=4.1cm]{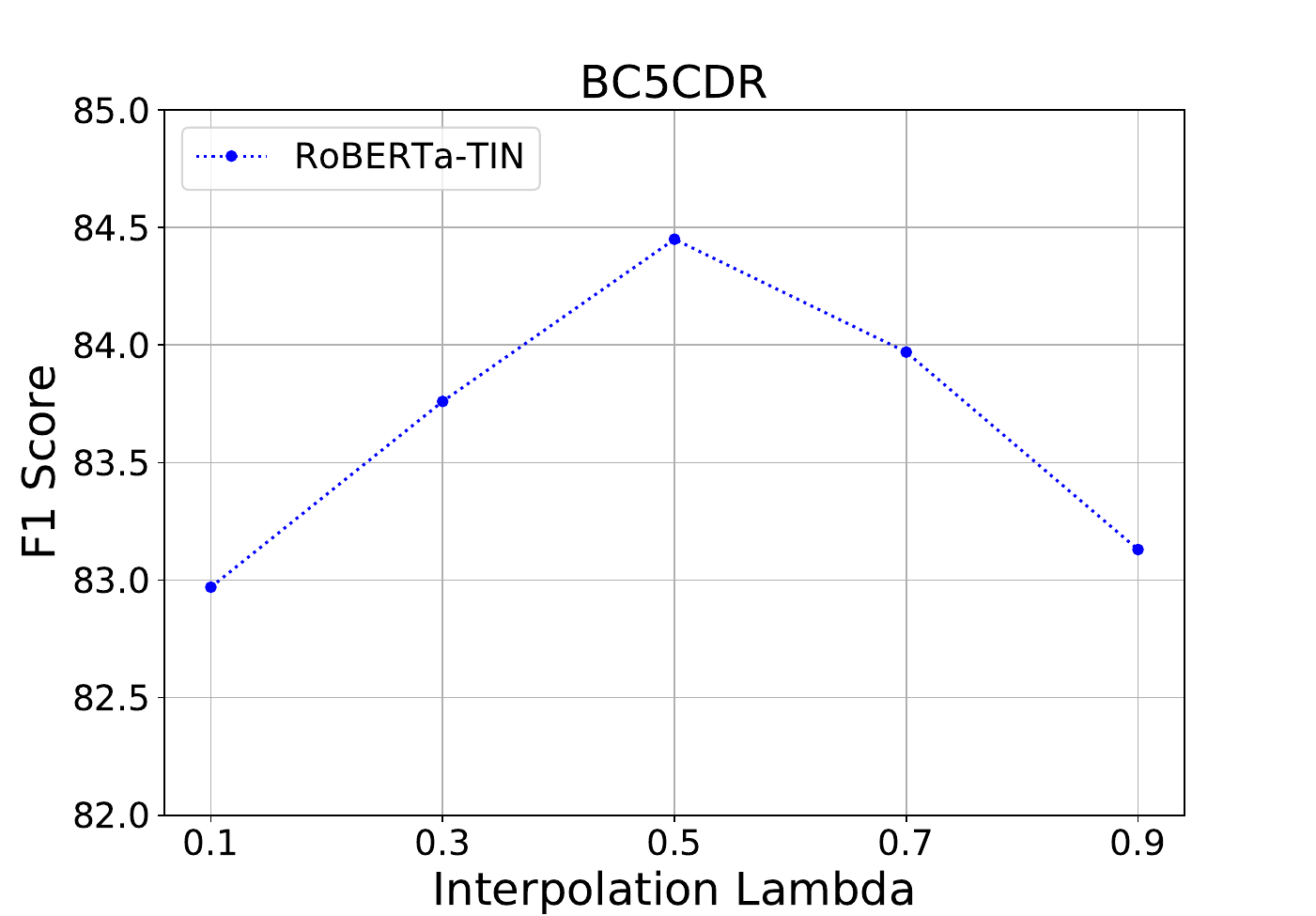}}
  \vspace{-0.1cm}
  \centerline{\small (b) RoBERTa+TIN+CRF}
\end{minipage}
\caption{Performance comparison (F1-score) of entity+trigger baselines on 20\% training dataset of BC5CDR with different interpolation weight $\lambda$.}
\label{fig:interpolation}
\end{figure}

\smallskip
\noindent
\textbf{Sensitivity Analysis of interpolation hyper-parameter ($\lambda$).}
In Sec~\ref{ssec:tin}, we linearly interpolated two different sources of knowledge by weight $\lambda$ 0.5.
To show how the weight $\lambda$ affects the performance, we conduct an ablation study on different $\lambda$ distribution.
As we can see from Figure.~\ref{fig:interpolation}, the framework achieves the highest performance when $\lambda$ is set to 0.5. It supports that the model achieves the best when we interpolate the entity and trigger knowledge in equal.

\begin{figure}[t!]
\hspace{0.1cm}
\begin{minipage}{0.23\textwidth}
  \centerline{\includegraphics[width=4.2cm]{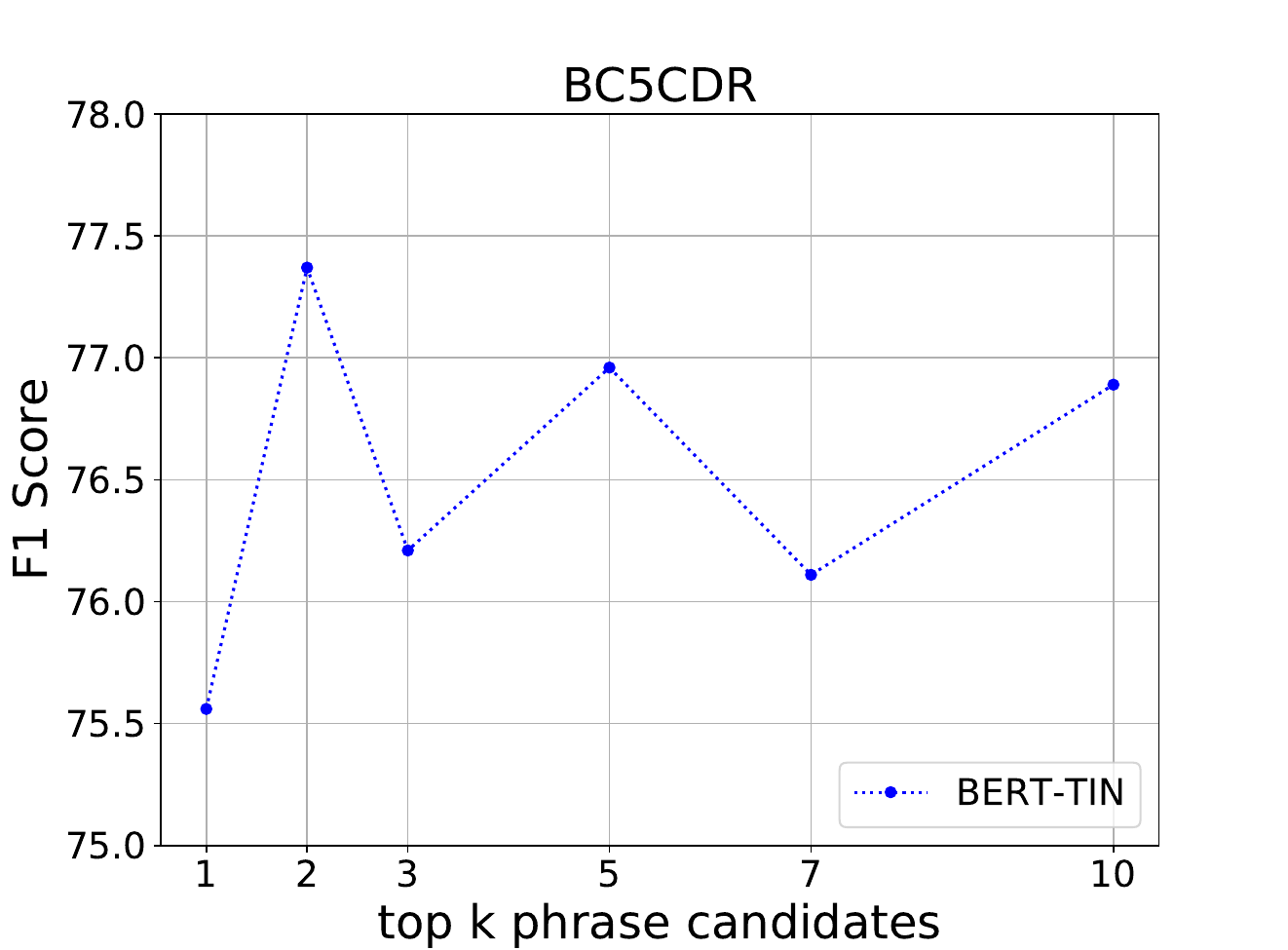}}
  \vspace{-0.1cm}
  \centerline{\small (a) BERT+TIN+CRF}
\end{minipage}
\begin{minipage}{0.23\textwidth}
  \centerline{\includegraphics[width=4.2cm]{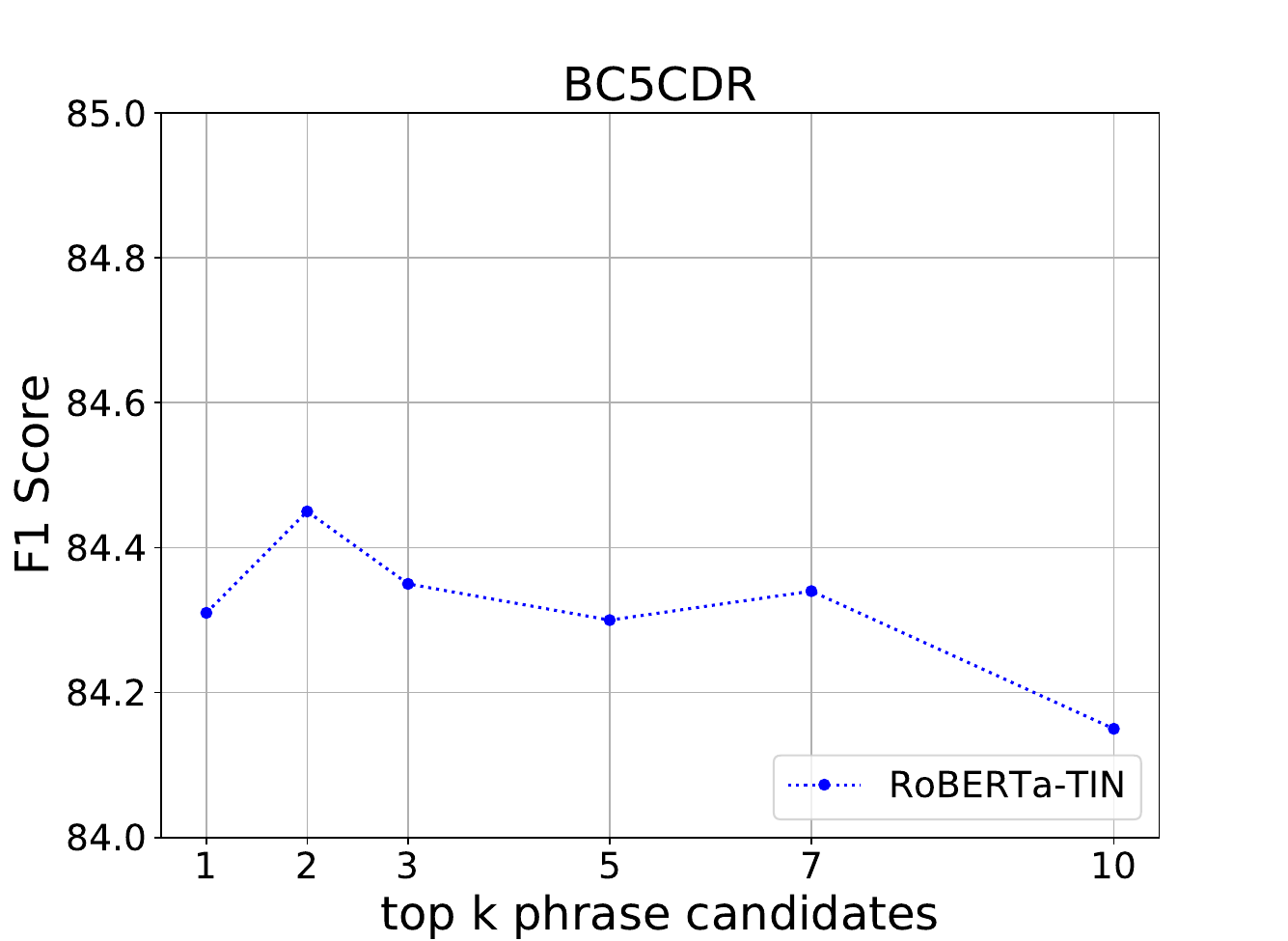}}
  \vspace{-0.1cm}
  \centerline{\small (b) RoBERTa+TIN+CRF}
\end{minipage}
\caption{Performance comparison (F1-score) of entity+trigger baselines on 20\% training dataset of BC5CDR with different number of triggers $k$.}
\label{fig:topk}
\end{figure}

\paragraph{Effect of number of triggers}
In Sec.~\ref{ssec:soc}, we pick the top $k$ candidate phrases with the highest importance score as the entity triggers after obtaining the importance score for all phrase candidates. 
For our main experiment, we use top 2 candidate phrases (see Table~\ref{tab:overall}). 
To show how the number of triggers affects the performance, we conduct an ablation study on model performance by different $k$. As we can see from Figure.~\ref{fig:topk}, the framework achieves the highest performance when we use top 2 phrase candidates as triggers.

\subsection{Related Works}
\smallskip
\noindent
\textbf{NER with Additional Supervision}
Previous and recent research has shown that encoding syntactic information into NER models compensate for the lack of labeled data \citep{syntactic_Tian2020ImprovingBN}. The improvement is consistent across word embedding based encoding (e.g. biLSTM) as well as unsupervised language model based encoding (e.g. BioBERT~\cite{lee2020biobert}) of the given text. Typically, the external information that is encoded include POS labels, syntactic constituents, and dependency relations \citep{ syntactic_nie-etal-2020-improving, syntactic_Tian2020ImprovingBN}. The general mechanism to include linguistic information into NER model is to represent them using word vectors and then concatenate those representations with the original text representation. This approach fails to identify the importance of different types of syntactic information. Recently, \citet{syntactic_Tian2020ImprovingBN} and \citet{syntactic_nie-etal-2020-improving} both showed that key-value memory network (KVMN)  \citep{kvmn_miller2016key} are effective in capturing importance of linguistic information arising from different sources. KVMN has been shown to be effective in leveraging extra information, such as knowledge base entities, to improve question answering tasks. Before applying KVMN, contextual information about a token is encoded as the key and syntactic information are encoded as values. Finally, weights over the values are computed using the keys to obtain a representation of the values and concatenate it with the context features. Our approach uses token level features extracted by an explanation generation model, but later train to be able to pick-up those explanations directly from the text at inference time. 

\begin{figure*}[t!]
\centerline{\includegraphics[width=16cm]{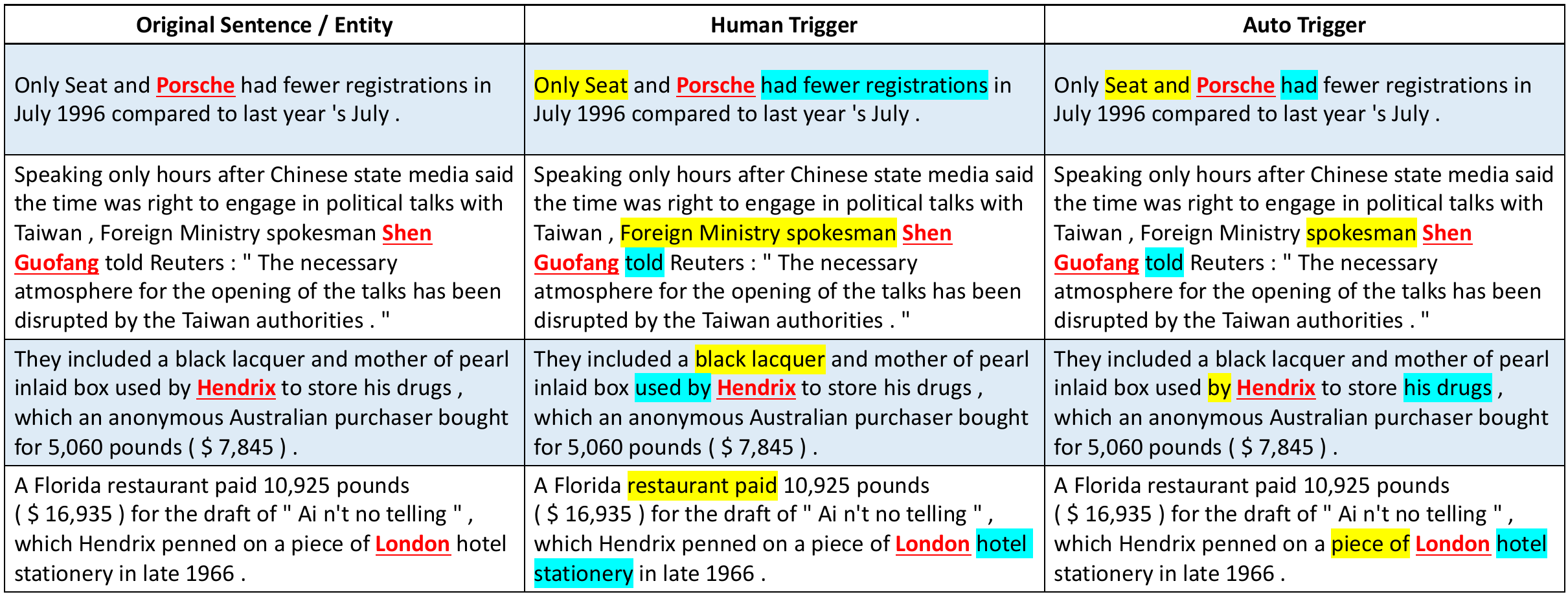}}
\caption{Case examples of \texttt{auto} trigger and \texttt{human} trigger. Entities are \textbf{bold} and \underline{underlined} with red color, and its triggers are highlighted. Different triggers are color-coded.}
\label{fig:triganalysis}
\end{figure*}


\smallskip
\noindent
\textbf{Limited Training Data for NER.}\quad 
The simplest way to approach the problem of limited data for NER is to use dictionary based weak supervision. An entity dictionary is used to retrieves unlabeled sentences from a corpus and weakly label them to create additional noisy data. This approach suffers from low recall as the training data covers a limited number of entities. The models tend to bias towards the surface form of the entities it has observed in the dictionary. There has also been approaches to retrieve sentences from a large corpus that are similar to sentences in the low-resource corpus to enrich it. These self-training approaches have been shown to be effective both in extremely limited data \citep{foley2018named, sarwar2018term} as well as limited data scenario \citep{du2020self}. Even though these data enhancement approaches explore a corpus to find related data cases, they do not exploit the explanation-based signals that is available within the limited data. 

\smallskip
\noindent
\textbf{Learning from Explanations.}\quad 
Recent works on Explainable AI are primarily focused on debugging the black box models by probing internal representations~\citep{ Adi2017FinegrainedAO, Conneau2018WhatYC},  testing model behavior using challenge sets~\citep{McCoy2019RightFT, Gardner2020EvaluatingML, Ribeiro2020BeyondAB}, or analyzing an impact of input examples by input perturbations or influence function looking at input examples~\citep{Ribeiro2016WhySI, Koh2017UnderstandingBP}. However, for an explanation of the model to be effective, it must provide not only the reasons for the model's prediction but also suggestions for corresponding actions in order to achieve an objective. 
Efforts to cope with this issue by incorporating human explanations into the model are called Explanation-based learning~\citep{DeJong2004ExplanationbasedLA}. 
These works are aiming to exploit generalized explanations for drawing inferences from unlabeled data while maintaining model transparency. Most prior works on explanation-based learning are mainly focused on facilitating logical rules as an explanation. They use such rules to create weak supervision~\citep{Ratner2017SnorkelRT} and regularize posterior~\citep{Hu2016HarnessingDN, Hu2017TowardCG}. Another form of explanations can be specific words in the sentence which aligns to our work. Notable work in this line asks annotators to highlight important words, then learn a generative model over parameters given these rationales~\citep{Zaidan2008ModelingAA}.

\begin{table*}[t]
	\centering
	\small
	\scalebox{1}{
		\begin{tabular}{ccccc}
			\toprule   
			\multirow{2}{*}{\textbf{Dataset}} & 
			\multirow{2}{*}{\textbf{Entity Type}} &
            \multicolumn{1}{c}{Original $\mathcal{D}_{L}$} &
            \multicolumn{2}{c}{Crowd-sourced trigger $\mathcal{D}_{HT}$} \\
            \cmidrule(lr){3-3} \cmidrule(lr){4-5} & &
			\multicolumn{1}{c}{\# of Entities} & 
			\multicolumn{1}{c}{\# of Entities} & 
			\multicolumn{1}{c}{\# of Human Triggers} 
			\\ 
			\midrule
			\textsc{CONLL 2003} & \textsc{PER} & 6,599 & 1,608 & 3,445 \\
			& \textsc{ORG} & 6,320 & 958 & 1,970  \\  
			& \textsc{MISC} & 3,437 & 787 & 2,057  \\ 
		    & \textsc{LOC} & 7,139 & 1,781 & 3,456 \\ 
		    \midrule
			& \textbf{Total} & 23,495 & 5,134 & 10,938 \\ 
			\midrule\midrule
			\textsc{BC5CDR} & \textsc{Disease} & 4,181 & 906 & 2,130 \\
			& \textsc{Chemical} & 5,202 & 1,085 & 1,640   \\ 
			\midrule
			& \textbf{Total} & 9,383 & 1,991 & 3,770 \\ 
			\midrule\midrule
			\textsc{JNLPBA} & \textsc{Protein} & 27,802 & -  & -\\
			& \textsc{DNA} & 8,480 & - &  - \\ 
		    & \textsc{RNA} & 843 & - & - \\ 
		    & \textsc{Cell Line} & 3,429 &-  & - \\ 
		    & \textsc{Cell Type} & 6,191 & - & - \\ 
		    \midrule
			& \textbf{Total} & 46,745 & - & - \\ 
			\bottomrule
		\end{tabular}
	}
	
	\caption{Train data statistics.}
	\label{tab:dataset}
\end{table*}

\end{document}